\pdfoutput=1

\documentclass[11pt]{article}

\usepackage[]{EMNLP2023}

\usepackage{times}
\usepackage{tabu}
\usepackage{latexsym}
\usepackage{stmaryrd}
\usepackage{amsmath}
\usepackage{multirow}
\usepackage{bbm}
\usepackage{graphicx}
\usepackage{amssymb}
\usepackage{booktabs}
\usepackage{adjustbox}
\usepackage{xspace}
\usepackage{algpseudocode}
\usepackage{algorithm}
\usepackage{graphicx}
\usepackage{caption}
\usepackage{subcaption}
\usepackage{tabularx}
\usepackage{colortbl}
\usepackage{xcolor}
\usepackage{tablefootnote}
\usepackage{enumitem}  
\usepackage{lipsum}
\usepackage{xspace}
\usepackage{bbding}
\usepackage{pifont}
\usepackage{arydshln}
\usepackage{array}
\newcommand{\PreserveBackslash}[1]{\let\temp=\\#1\let\\=\temp}
\newcolumntype{C}[1]{>{\PreserveBackslash\centering}p{#1}}
\newcolumntype{R}[1]{>{\PreserveBackslash\raggedleft}p{#1}}
\newcolumntype{L}[1]{>{\PreserveBackslash\raggedright}p{#1}}

\usepackage{hyperref}

\def\xHyphenate#1#2\wholeString {\if#1$%
    \else\transform{#1}%
    \takeTheRest#2\ofTheString\fi}
\def\takeTheRest#1\ofTheString\fi
{\fi \xHyphenate#1\wholeString}
\def\transform#1{\url{#1}\hskip 0pt plus 1pt}

\usepackage[T1]{fontenc}

\usepackage[utf8]{inputenc}

\usepackage{microtype}

\newcommand{\datasetname}{IndoMMLU}
 
\newcommand{\ex}[1]{\textit{#1}\xspace} 
\newcommand{\gl}[1]{``#1''\xspace}

\definecolor{mygreen}{RGB}{217, 234, 211}
\definecolor{myred}{RGB}{244, 204, 204}

\newcommand{\ok}{\cellcolor{mygreen}}
\newcommand{\no}{\cellcolor{myred}}

\usepackage{inconsolata}

\setlist{topsep=1pt,itemsep=1pt,partopsep=1pt, parsep=1pt}

\title{Large Language Models Only Pass Primary School Exams in Indonesia: \\A Comprehensive Test on \datasetname{}}

\author{Fajri Koto$^{1}$ \qquad  Nurul Aisyah$^{2}$ \qquad Haonan Li$^{1}$  \qquad     Timothy Baldwin$^{1,3}$ \\ 
	$^{1}$Department Natural Language Processing, MBZUAI \\
        $^{2}$Quantic School of Business and Technology \\
        $^{3}$The University of Melbourne \\
	\texttt{\small fajri.koto@mbzuai.ac.ae, nurulaisyah.inc@gmail.com, \{haonan.li,timothy.baldwin\}@mbzuai.ac.ae 
	} \\
}

\begin{document}
\maketitle

\begin{abstract}


Although large language models (LLMs) are often pre-trained on  large-scale multilingual texts, their reasoning abilities and real-world knowledge are mainly evaluated based on English datasets. Assessing LLM capabilities beyond English is increasingly vital but hindered due to the lack of suitable datasets. In this work, we introduce \texttt{IndoMMLU}, the first multi-task language understanding benchmark for Indonesian culture and languages, which consists of questions from primary school to university entrance exams in Indonesia. By employing professional teachers, we obtain 14,981 questions across 64 tasks and education levels, with 46\% of the questions focusing on assessing proficiency in the Indonesian language and knowledge of nine local languages and cultures in Indonesia. Our empirical evaluations show that GPT-3.5 only manages to pass the Indonesian primary school level, with limited knowledge of local Indonesian languages and culture. Other smaller models such as BLOOMZ and Falcon perform at even lower levels.\footnote{Code and dataset can be found at \url{https://github.com/fajri91/IndoMMLU}}

\end{abstract}

\section{Introduction}

The evaluation of large language models (LLMs) has predominantly relied on English datasets to assess language proficiency \cite{wang-etal-2018-glue,baradaran2022survey}, reasoning abilities \cite{zellers-etal-2019-hellaswag,huang-etal-2019-cosmos,bisk2020piqa,talmor-etal-2019-commonsenseqa}, and real-world knowledge \cite{hendrycksmeasuring}. LLMs such as GPT-3.5 \cite{ouyang2022training}, Falcon \cite{penedo2023refinedweb}, and BLOOMZ \cite{muennighoff2022crosslingual}, however, are pre-trained on large-scale multilingual data, and thus it is critical to evaluate what knowledge they capture and their reasoning abilities in languages beyond English. 

School exams serve as a powerful means to assess the reasoning abilities and real-world knowledge of LLMs, given that these tests are meticulously designed by expert educators, drawing upon the principles of learning science. At various educational levels, school exams function as assessment tools, evaluating not only language proficiency but also higher-order cognitive skills such as comprehension, analytic abilities, and the application of real-world knowledge across diverse scenarios \cite{novak1988learning}.

\citet{hendrycksmeasuring} proposed \texttt{MMLU}, a massive multitask language understanding benchmark in English that is compiled from different exams, covering topics including US history, computer science, and high school subjects. Recent progresses on LLMs such as LLaMA \cite{touvron2023llama} and GPT--4 \cite{OpenAI2023GPT4TR} use \texttt{MMLU} as one of the evaluation datasets. In the GPT-4 technical report, automatic evaluation is further extended to encompass various standardized exams, including SAT, GRE, and bar exams.

\begin{figure}[t]
    \centering
    \includegraphics[width=\linewidth]{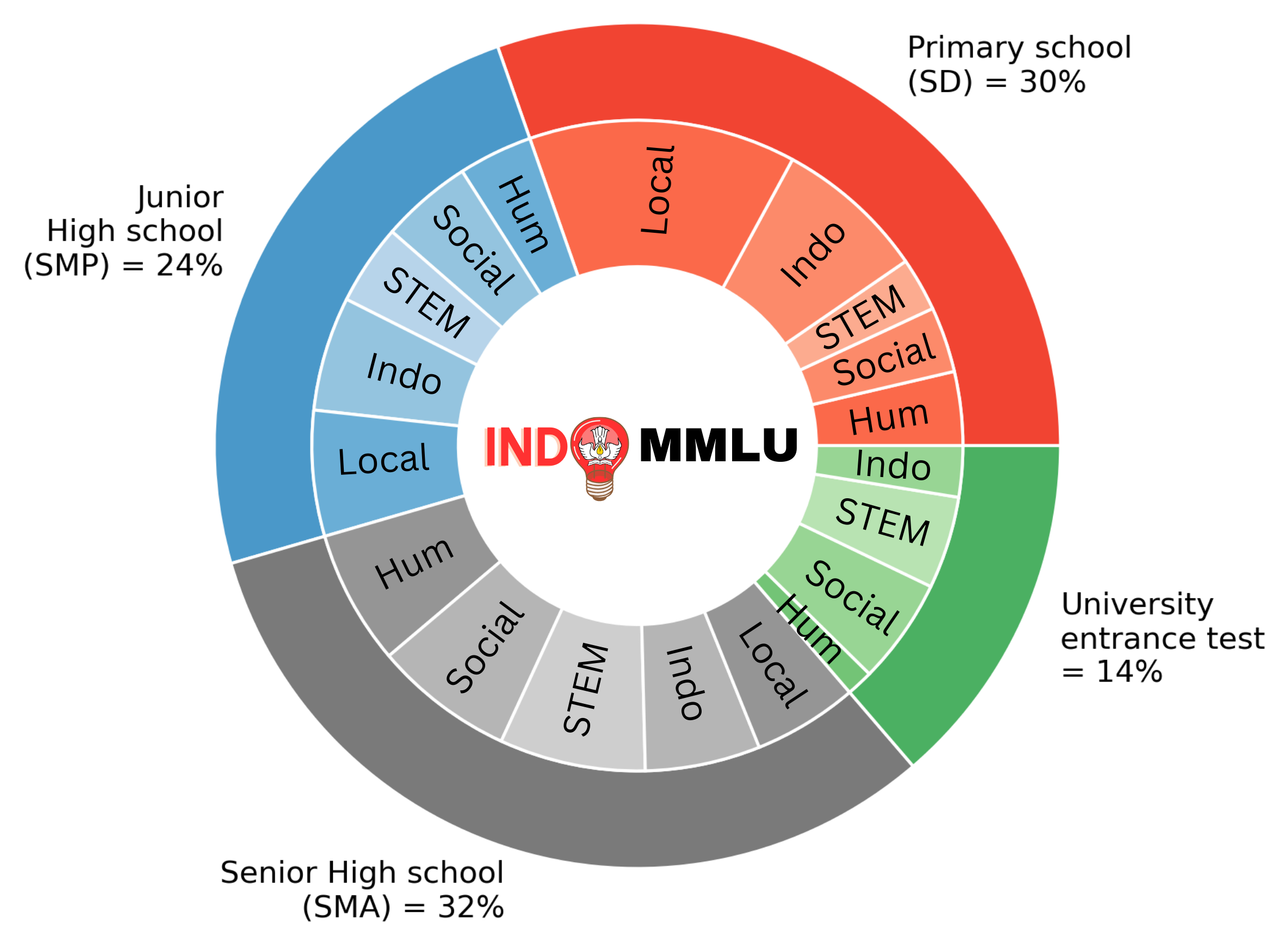} 
    \caption{Distribution of subject areas and education levels in \texttt{IndoMMLU}. ``Hum'', ``Social'', ``Indo'', and ``Local'' refer to Humanities, Social Science, Indonesian Language, and Local Languages and Cultures, respectively. }
    \label{fig:overview}
\end{figure}

While there has been a plethora of work on LLM evaluation for English \cite{OpenAI2023GPT4TR,katz2023gpt,choi2023chatgpt,ryznar2023exams,chalkidis2023chatgpt}, there has been comparatively little work in other languages \cite{li2023cmmlu,sengupta2023jais}. Recent work by \citet{OpenAI2023GPT4TR} evaluated GPT-4 using a translated version of \texttt{MMLU}, and reported strong performance. While encouraging, using translations of English evaluation datasets has serious shortcomings, including translation noise, a complete lack of content that is sensitized to the local language/culture (esp.\ as most English evaluation datasets are highly US centric), and conversely, the existence of content that is irrelevant to the local language/culture (e.g.\ questions relating to US law or customs) and incongruent with the language-specific evaluation \cite{liu2023multilingual}.

In this paper, we ask professional teachers (of Indonesian nationality) to collect exam questions from various educational levels in Indonesian schools (i.e.~primary school, junior high school, senior high school, and university). We categorize the collected questions into different subject areas, including: (1) STEM (Science, Technology, Engineering, and Mathematics); (2) Social Science; (3) Humanities; (4) Indonesian Language; and (5) Local Languages and Cultures. Figure~\ref{fig:overview} presents an overview of the distribution of the resulting dataset, \texttt{IndoMMLU}, across different subject areas and education levels. It is worth mentioning that 21\% of the questions specifically focus on the Indonesian language, and 25\% encompass nine distinct local languages and cultures that are specific to Indonesia.

Our contributions can be summarized as follows:
\begin{itemize}
    \item We introduce the first Indonesian \texttt{MMLU} dataset, namely \texttt{IndoMMLU}, which comprises 64 tasks across different subject areas and education levels in Indonesia. 
    \item Our dataset includes exam questions from school grades 1 to 12, as well as university entrance exams. This comprehensive coverage allows us to perform fine-grained assessment of the Indonesian language proficiency of existing LLMs.
    \item Approximately 25\% of our data encompasses nine distinct local languages and cultures in Indonesia, namely Lampungic (\texttt{ljp}), Balinese (\texttt{ban}), Makassarese (\texttt{mak}), Banjarese (\texttt{bjn}), Madurese (\texttt{mad}), Sundanese (\texttt{sun}), Javanese (\texttt{jav}), Dayak Ngaju (\texttt{nij}), and Minangkabau.\footnote{For Minangkabau culture, the Indonesian language is used in teaching and exams.}
    These questions are not only in under-represented languages but also incorporate specific cultural content, such as art, poetry, and daily life. For Lampungic (\texttt{ljp}) and Makassarese (\texttt{mak}) in particular, this is the very first NLP resource to be released.
    \item We evaluate various multilingual LLMs, including GPT-3.5 \cite{ouyang2022training}, XGLM \cite{lin2021few}, Falcon \cite{penedo2023refinedweb}, BLOOMZ \cite{muennighoff2022crosslingual}, mT0 \cite{muennighoff2022crosslingual}, LLaMA \cite{touvron2023llama}, and Bactrian-X \cite{li2023bactrian}, across different model sizes. We find that only GPT-3.5 passes the highest primary school level exam, and no models demonstrate familiarity with local Indonesian languages and culture.
\end{itemize}

\section{Related Work}
 
\paragraph{Evaluating Large Language Models} Various benchmarks have been released to evaluate English pre-trained LMs \cite{devlin-etal-2019-bert,conneau-etal-2020-unsupervised}. Early benchmarks such as \texttt{GLUE} \cite{wang-etal-2018-glue} and \texttt{SuperGLUE} \cite{wang2019superglue} consist of various natural language understanding (NLU) tasks of different types with varying training data sizes. \texttt{XGLUE} \cite{liang-etal-2020-xglue}, \texttt{XTREME} \cite{hu2020xtreme}, and \texttt{XTREME-R} \cite{ruder-etal-2021-xtreme} serve as multilingual benchmarks of more than 20 languages. For natural language generation (NLG), the \texttt{GEM} benchmark \cite{gehrmann-etal-2021-gem} is a collection of machine translation, summarization, and generated descriptions in many languages. 

As LLMs have become larger in size and improved over the standard benchmarks, there has been a shift in evaluation practice to focus on reasoning abilities \cite{zellers-etal-2019-hellaswag,huang-etal-2019-cosmos,bisk2020piqa,talmor-etal-2019-commonsenseqa,koto-etal-2022-cloze}, and real-world knowledge \cite{hendrycksmeasuring}. In GPT-4 \cite{OpenAI2023GPT4TR}, for instance, commonsense reasoning is evaluated using \texttt{HellaSwag} \cite{zellers-etal-2019-hellaswag} and \texttt{WinoGrande} \cite{sakaguchi2021winogrande}, while real-world knowledge is evaluated based on school exams including \texttt{MMLU} \cite{hendrycksmeasuring}, \texttt{ARC} \cite{clark2018think}, and \texttt{GSM-8K} \cite{cobbe2021training}. Similarly, LLaMA \cite{touvron2023llama} was evaluated using school exam problems, in addition to closed-book question answering \cite{kwiatkowski-etal-2019-natural,joshi-etal-2017-triviaqa} and the \texttt{RACE} reading comprehension benchmark \cite{lai-etal-2017-race}.

\paragraph{Indonesian Pre-trained Language Models and Benchmarks} 
Several monolingual pre-trained language models have been released for Indonesian, including IndoBERT \cite{koto-etal-2020-indolem,wilie-etal-2020-indonlu}, IndoBERTweet \cite{koto-etal-2021-indobertweet}, and IndoBART \cite{cahyawijaya-etal-2021-indonlg}. These models have been evaluated on NLU (e.g.~IndoLEM and IndoNLU) and NLG (e.g.~IndoNLG) benchmarks. Component tasks include sentiment analysis \cite{koto2017inset,purwarianti2019improving}, emotion classification \cite{saputri2018emotion}, hate speech detection, summarization \cite{koto-etal-2020-liputan6,koto-etal-2022-lipkey}, and translation \cite{cahyawijaya-etal-2021-indonlg,koto-koto-2020-towards}.

In contemporaneous work, \citet{cahyawijaya2022nusacrowd} evaluated several LLMs using existing Indonesian datasets. This collection includes the recent \texttt{NusaX} dataset \cite{winata-etal-2023-nusax}, which is a parallel sentiment analysis dataset in 10 Indonesian local languages, created through human translation. The collection also includes several question-answering datasets, such as \texttt{FactQA} \cite{purwarianti2007machine}, \texttt{IDK-MRC} \cite{putri-oh-2022-idk}, and \texttt{TyDiQA} \cite{clark-etal-2020-tydi}, over news and Wikipedia documents. \texttt{IndoMMLU} is different in that it explicitly evaluates reasoning, language, and cultural abilities in a fine-grained manner from the perspective of education science.

\begin{figure}[t]
    \centering
    \includegraphics[width=\linewidth]{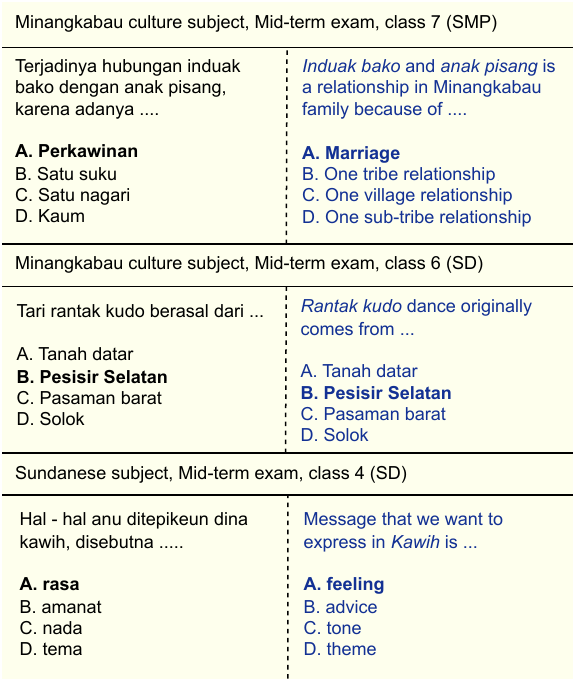} 
    \caption{The first question  focuses on the family relationship between \ex{anak pisang} \gl{children} and \ex{induak bako} \gl{aunt on the father's side}. Both terms are commonly used in Minangkabau but not in the Indonesian language. The second and third questions pertain to traditional art. \ex{Kawih} in the third question means a song set to a distinctive beat in Sundanese culture. \textbf{Left} is the original text and \textbf{right} is the English translation for illustrative purposes. The bold options are the correct answer keys.}
    \label{fig:ex1}
\end{figure}

\begin{figure}[t]
    \centering
    \includegraphics[width=\linewidth]{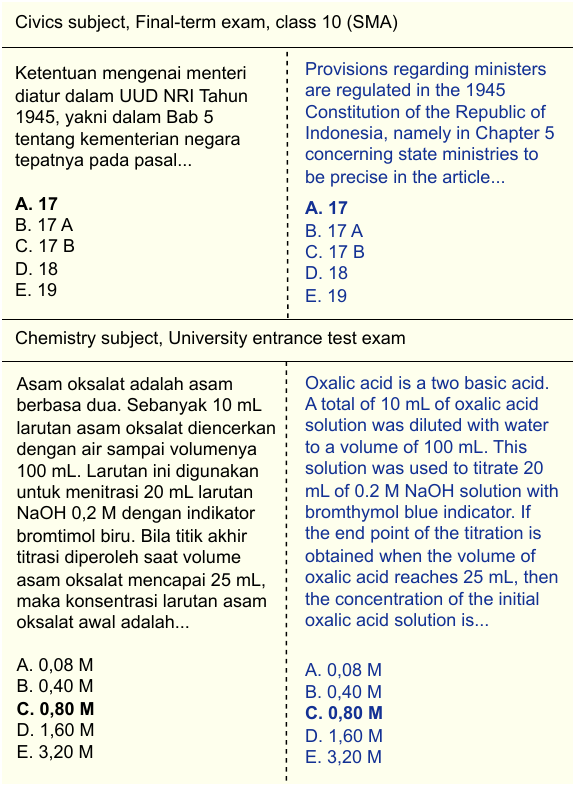} 
    \caption{Examples of civics and chemistry exam questions. \textbf{Left} is the original text and \textbf{right} is the English translation for illustrative purposes. The bolded options are the answer keys.}
    \label{fig:ex2}
\end{figure}

\section{{IndoMMLU}}
\label{sec:data_construct}

\begin{table}[t]
    \centering
    \resizebox{\linewidth}{!}{
        \begin{tabular}{lrr}
        \toprule
        \textbf{Group} & \textbf{Question} & \textbf{Answer} \\        
        \midrule
        Primary school & 99.6 & 65.5 \\
        Junior high school & 188.3 & 105.9 \\
        Senior high school & 167.7 & 130.3 \\
        University Entrance Test & 204.9 & 186.2 \\
        \midrule
        STEM & 133.7 & 102.4 \\
        Social science & 136.2 & 131.9 \\
        Humanities & 113.2 & 104.4 \\
        Local languages and cultures & 88.4 & 68.0 \\
        Indonesian language & 307.4 & 161.8 \\
        \bottomrule
        \end{tabular}
    }
    \caption{Average question and answer length (in characters) for each education group and subject area.}
    \label{tab:stat}
\end{table}

\begin{table}[t]
    \centering
    \resizebox{\linewidth}{!}{
        \begin{tabular}{L{3.2cm}L{6cm}}
        \toprule
        \textbf{Group} & \textbf{Subjects} \\        
        \midrule
            STEM & Chemistry (SMA, UE), Biology (SMA, UE), Physics (SMA, UE), Science (SD, SMP)  \\ \midrule
            Social science & Geography (SMA, UE), Sociology (SMA, UE), Economy (SMA, UE), Civics education (SD, SMP, SMA), Social science (SD, SMP) \\ \midrule
            Humanities & History (SMA, UE), Art (SD, SMP, SMA), Sports (SD, SMP, SMA), Islam religion (SD, SMP, SMA), Christian religion (SD, SMP, SMA), Hindu religion (SD, SMP, SMA)  \\ \midrule
            Local languages and cultures & Lampungic (SD, SMP, SMA), Balinese (SD, SMP, SMA), Makassarese (SD, SMP, SMA), Banjarese (SD, SMP, SMA), Madurese (SD, SMP, SMA), Minangkabau culture (SD, SMP), Dayak Ngaju (SD), Sundanese (SD, SMP, SMA), Javanese  (SD, SMP, SMA)\\ \midrule
            Indonesian language & Indonesian language  (SD, SMP, SMA, UE)\\
        \bottomrule
        \end{tabular}
    }
    \caption{Subject areas in \texttt{IndoMMLU}. ``SD'', ``SMP'', ``SMA'', ``UE'' indicate  that questions in the subject are are available in primary school, junior high school, senior high school, and university entrance exams, respectively.}
    \label{tab:group}
\end{table}

\texttt{IndoMMLU} is a multiple-choice problem set in 64 subjects from different education levels, following the format of English \texttt{MMLU} (see Figure~\ref{fig:ex1} and  Figure~\ref{fig:ex2}). \texttt{IndoMMLU}, however, is based on the Indonesian education curriculum, and has more fine-grained education levels than \texttt{MMLU}. 

In Indonesia's curriculum, schools are categorized into three levels: (1) six years of primary school (\ex{Sekolah Dasar} = ``SD''), (2) three years of junior high school (\ex{Sekolah Menengah Pertama} = ``SMP''), and (3) three years of senior high school (\ex{Sekolah Menengah Atas} = ``SMA''). At primary school, pupils in all grades are taught the Indonesian language, civics, mathematics, art, sports, and religion. From grade 4 to 6 and in junior high school, pupils additionally learn a foreign language, a local language/culture, science, and social science.\footnote{In a recent curriculum change, science and social science have been added from grade 3.} In senior high school, pupils study more specialized natural science and social science subjects, including physics, chemistry, biology, geography, sociology, economy, and history. In \texttt{IndoMMLU}, we explicitly exclude mathematics because the questions typically consist primarily of symbols with little language content, and there are existing datasets for mathematical reasoning such as \texttt{GSM-8K} \cite{cobbe2021training} and \texttt{NumGLUE} \cite{mishra-etal-2022-numglue}.

The local language/culture subjects vary across provinces in Indonesia and depend on the local government policy. For example, in West Sumatra, Minangkabau culture is taught using the Indonesian language, while in West Java, pupils are exposed to the Sundanese language and culture. Figure~\ref{fig:ex1} illustrates two exam questions for Minangkabau culture, and one exam question for Sundanese.

\subsection{Data Construction}
We asked seven professional teachers with at least a bachelor's degree in education to gather publicly-available school exam questions in Indonesia from web sources.\footnote{The seven teachers were selected from 70 applicants.} They were tasked with gathering problems for specific subject areas and educational levels, as well as metadata such as the source (i.e.\ URL of the source document), school level, class level, question, multiple-choice options, and the correct answer key. 
We instructed the teachers to only include exams that had accompanying answer keys, and to exclude problems that contained images. Additionally, we organized an 1-hour workshop to discuss the data collection procedure with all the teachers, addressing any questions or concerns they had. All teachers are paid competitively, higher than the Indonesian average monthly wage.\footnote{The work for a single teacher was equivalent to a 5-day full-time job.}

\subsection{Quality Control}

To ensure the accuracy of the data entry process,  we randomly checked questions collected by each teacher. We manually verified the questions, multiple-choice options, and the corresponding answer keys based on the given URL, and found that each teacher conducted the work accurately. We also additionally performed automatic filtering to remove repetitive questions, and remove questions that have no answer key. 

\subsection{Data Statistics}
After data cleansing, we obtained a total of 14,981 questions, distributed over school levels and subjects as detailed in Figure~\ref{fig:overview}; the details of each subject area are in Table~\ref{tab:group} and the Appendix. \texttt{IndoMMLU} consists of 30\% primary school, 24\% junior high school, 32\% senior high school, and 14\% university entrance exam questions.  Table~\ref{tab:stat} shows the average question length for each education level and subject area. We can observe that primary school questions tend to be shorter and university entrance exam questions are longer. Indonesian language questions have the highest average length, while local languages and culture questions are around 88 characters on average.

\section{{Experiments}}

\begin{figure}[t]
    \centering
    \includegraphics[width=0.9\linewidth]{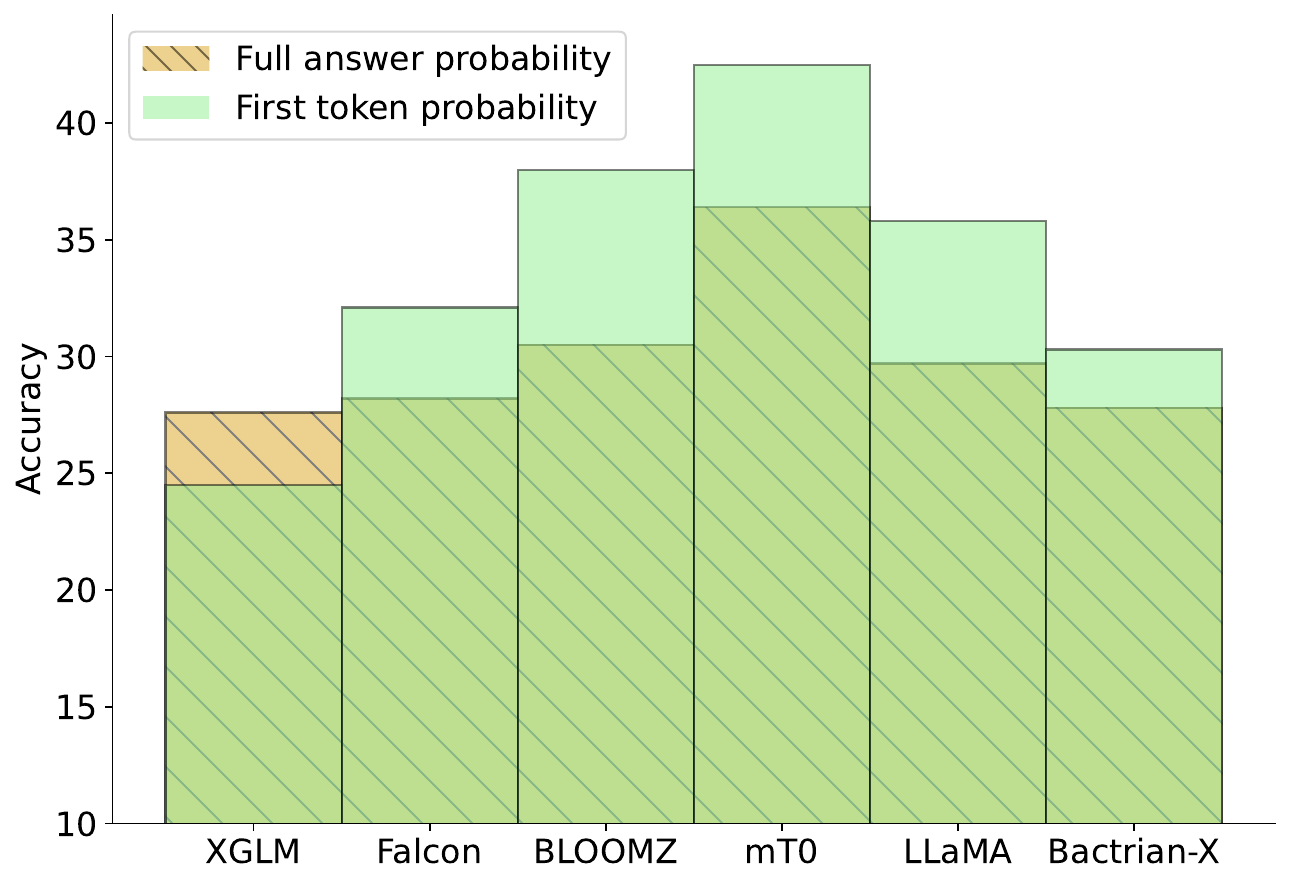} 
    \caption{LLM performance (\% accuracy) based on: (1) the probability of the full generated answer; and (2) the probability of the first token in the generated answer.}
    \label{fig:benchmark}
\end{figure}

\begin{table*}[t]
    \centering
    \resizebox{0.95\linewidth}{!}{
        \begin{tabular}{lC{1cm}C{1.5cm}C{2cm}C{2cm}C{3cm}C{1.5cm}}
        \toprule
       \multirow{2}{*}{\textbf{Model (\#parameters)}}& \multirow{2}{*}{\textbf{STEM}} & \textbf{Social} & \multirow{2}{*}{\textbf{Humanities}} & \textbf{Indonesian} & \textbf{Local languages}  & \multirow{2}{*}{\textbf{Average}} \\
       & & \textbf{Science} & & \textbf{Language} & \textbf{and Cultures} & \\
       \midrule
        Random & 21.9 & 23.4 & 23.5 & 24.4 & 26.6 & 24.4 \\
        GPT-3.5 (175B) & \textbf{54.3} & \textbf{62.5} & \textbf{64.0} & \textbf{62.2} & {39.3} & \textbf{53.2} \\
       \hdashline
        XGLM (564M) & 22.1 & 23.0 & 25.6 & 25.6 & 27.5 & 25.2 \\
        XGLM (1.7B) & 20.9 & 23.0 & 24.6 & 24.8 & 26.6 & 24.4 \\
        XGLM (2.9B) & 22.9 & 23.2 & 25.4 & 26.3 & 27.2 & 25.2 \\
        XGLM (4.5B) & 21.8 & 23.1 & 25.6 & 25.8 & 27.1 & 25.0 \\
        XGLM (7.5B) & 22.7 & 21.7 & 23.6 & 24.5 & 27.5 & 24.5 \\
        \hdashline
        Falcon (7B) & 22.1 & 22.9 & 25.5 & 25.7 & 27.5 & 25.1 \\
        Falcon (40B) & 30.2 & 34.8 & 34.8 & 34.9 & 29.2 & 32.1 \\
        \hdashline
        BLOOMZ (560M) & 22.9 & 23.6 & 23.2 & 24.2 & 25.1 & 24.0 \\
        BLOOMZ (1.1B) & 20.4 & 21.4 & 21.1 & 23.5 & 24.7 & 22.4 \\
        BLOOMZ (1.7B) & 31.5 & 39.3 & 38.3 & 42.8 & 29.4 & 34.4 \\
        BLOOMZ (3B) & 33.5 & 44.5 & 39.7 & 46.7 & 29.8 & 36.4 \\
        BLOOMZ (7.1B) & 37.1 & 46.7 & 44.0 & 49.1 & 28.2 & 38.0 \\
        \hdashline
        mT0$_\text{small}$ (300M) & 21.8 & 21.4 & 25.7 & 25.1 & 27.6 & 24.9 \\
        mT0$_\text{base}$ (580M) & 22.6 & 22.6 & 25.7 & 25.6 & 26.9 & 25.0 \\
        mT0$_\text{large}$ (1.2B) & 22.0 & 23.4 & 25.1 & 27.3 & 27.6 & 25.2 \\
        mT0$_\text{xl}$ (3.7B) & 31.4 & 42.9 & 41.0 & 47.8 & 35.7 & 38.2 \\
        mT0$_\text{xxl}$ (13B) & 33.5 & 46.2 & 47.9 & 52.6 & \textbf{39.6} & 42.5 \\
        \hdashline
        LLaMA (7B) & 22.8 & 23.1 & 25.1 & 26.7 & 27.6 & 25.3 \\
        LLaMA (13B) & 24.1 & 23.0 & 24.4 & 29.5 & 26.7 & 25.3 \\
        LLaMA (30B) & 25.4 & 23.5 & 25.9 & 28.4 & 28.7 & 26.5 \\
        LLaMA (65B) & 33.0 & 37.7 & 40.8 & 41.4 & 32.1 & 35.8 \\
        \hdashline
        Bactrian-X-LLaMA (7B) & 23.3 & 24.0 & 26.0 & 26.1 & 27.5 & 25.7 \\
        Bactrian-X-LLaMA (13B) & 28.3 & 29.9 & 32.8 & 35.2 & 29.2 & 30.3 \\

        \bottomrule
        \end{tabular}
    }
    \caption{Zero-shot performance (\% accuracy) of LLMs, combined across education levels. ``Average'' means the average across all subject areas in \texttt{IndoMMLU}.}
    \label{tab:result}
\end{table*}

\subsection{Set-Up}
We evaluate 24 multilingual LLMs of different sizes in zero-shot and few-shot settings. This includes GPT-3.5 \cite{ouyang2022training},  XGLM \cite{lin2021few}, Falcon \cite{penedo2023refinedweb}, BLOOMZ \cite{muennighoff2022crosslingual}, mT0 \cite{muennighoff2022crosslingual}, LLaMA \cite{touvron2023llama} and Bactrian-X \cite{li2023bactrian}.\footnote{At the time this research was carried out, we did not have access to the GPT-4 API, and thus leave it to future work.} We add a simple prompt in the Indonesian language \ex{Ini adalah soal [subject] untuk [level]. Pilihlah salah satu jawaban yang dianggap benar!} \gl{This is a [subject] question for [level]. Please choose the correct answer!} prior to the question and multiple-choice options.

For closed-source models, we evaluate questions by comparing the first generated tokens (e.g., \ex{A},  \ex{B}, \ex{C}) and the answer key using a regular expression.\footnote{In cases where there is no match, we assign a random answer.} For open-sourced models, we benchmark two strategies. Given a question and the corresponding multiple-choice options, we calculate: (1) the probability of the full generated answer; and (2) the probability of the first token in the generated answer. For the first, we select the answer with the highest normalized log likelihood, and for the second, we simply select the key token (e.g., \ex{C}) with the highest probability among all possible keys.

\subsection{Results}

Figure~\ref{fig:benchmark} presents the zero-shot accuracy when using: (1) the full answer probability; and (2) the probability of the first token in the generated answer. Among the open-sourced language models (LLMs) including XGLM (7.5B), Falcon (40B), BLOOMZ (7.1B), mT0$_\text{xxl}$ (13B), LLaMA (65B), and Bactrian-X (13B), we find that estimating the answer based on the probability of the first token in the generated answer generally performs best, with the notable exception of XGLM. 
Thus, we report results under this configuration in the remaining sections; the full results for both settings can be found in the Appendix.

\paragraph{Results across all models} Table~\ref{tab:result} shows the average accuracy for each subject area across the 24 models. To compute the scores, we disregard the education level of the questions, and average scores based on the subject (e.g.~Biology), and finally combine the scores across all subject areas (e.g.~STEM). The random performance varies between 20\% to 27\% due to the differing number of multiple-choice options (i.e.~three to five).

Overall, we found that GPT-3.5 attains the highest overall accuracy, albeit low at 53.2\%. GPT-3.5 is also notably the highest in each subject area, except in local languages and culture subjects. Among the open-source models, we observe that mT0$_\text{xxl}$ (13B) achieves an average accuracy of 42.5\%. Falcon (40B) performs worse than mT0$_\text{xxl}$ (13B) and BLOOMZ (7B).


Performance based on model size varies, with smaller models such as BLOOMZ (7B) and mT0$_\texttt{xxl}$ being better than Falcon (40B) and LLaMA (65B). We suspect that this is due to the absence of the Indonesian language in Falcon and LLaMA's pre-training data. The poor performance of the 13B and 30B LLaMA models might imply that any ``emergent abilities'' of LLMs generally appear in the same or closely-related languages. This is further supported by Bactrian-X-LLaMA (13B), a LLaMA model fine-tuned on instruction datasets in 52 languages (including Indonesian), which obtain a $+$5\% average increment, compared to LLaMA (13B).

\begin{table}[t]
    \centering
    \resizebox{\linewidth}{!}{
        \begin{tabular}{lcccc}
        \toprule
        \textbf{Subject} & \textbf{SD} & \textbf{SMP} & \textbf{SMA} & \textbf{UE} \\
        \midrule
        Science & \ok 76.3 & \ok 67.8 & \no 52.8 & \no 43.7 \\
        Social science & \ok 84.6 & \ok 73.1 & \no 63.5 & \no 48.2 \\
        Indonesian language & \ok 74.7 & \no 61.8 & \no 55.1 & \no 42.3 \\
        Civics & \ok 64.6 & \ok 65.2 & \ok 65.4 & -- \\
        Sports & \ok 66.7 & \no 44.7 & \no 62.0 & -- \\
        Art & \ok 73.9 & \ok 71.2 & \no 58.7 & -- \\
        Islam religion & \ok 78.6 & \no 59.9 & \ok 67.7 & -- \\
        Christian religion & \ok 83.7 & \ok 77.6 & \no 62.0 & -- \\
        Hindu religion & \ok 66.7 & \no 62.0 & \no 55.1 & -- \\
        Sundanese & \no 50.0 & \no 45.1 & \no 37.9 & -- \\
        Javenese & \no 46.1 & \no 36.1 & \no 36.1 & -- \\
        Balinese & \no 32.2 & \no 38.5 & \no 36.1 & -- \\
        Makassarese & \no 33.7 & \no 48.8 & \no 38.3 & -- \\
        Banjarese & \no 50.0 & \no 44.4 & \no 28.6 & -- \\
        Lampungic & \no 40.0 & \no 30.0 & \no 33.3 & -- \\
        Madurese & \no 41.0 & \no 28.3 & \no 35.0 & -- \\
        Minangkabau culture & \no 38.0 & \no 52.2 & -- & -- \\
        Dayak Ngaju & \no 31.1 & -- & -- & -- \\
        \bottomrule
        \end{tabular}
    }
    \caption{GPT-3.5 performance (\% accuracy) across different education levels. ``SD'', ``SMP'', ``SMA'', ``UE'' indicate  primary school, junior high school, senior high school, and university entrance tests, respectively. Red indicates that the score is below the minimum passing threshold of 65, while green signifies a score at or above this minimum.}
    \label{tab:stat_edu}
\end{table}

\paragraph{Results across education levels}

As illustrated in Figure~\ref{fig:overview}, \texttt{IndoMMLU} includes detailed education level metadata, which enables us to gain a deeper understanding of the capabilities of LLMs in terms of human education levels. In the Indonesian context, the minimum passing score for exams varies across subjects and typically ranges between 65 and 70.\footnote{This refers to Curriculum 2013 in Indonesia.} By setting the passing score at 65, we assess GPT-3.5 over real-world knowledge capabilities, as shown in Table~\ref{tab:stat_edu}.  Green indicates that the model has successfully passed the subject, while red indicates it has failed. This reveals that GPT-3.5 generally performs well on primary school exams for general subjects, but exhibits a lack of understanding of local languages and culture. In subjects that require less analytical thinking, such as civics and religion, GPT-3.5 tends to achieve higher scores in high school exams.

\begin{figure}[t]
    \centering
    \includegraphics[width=0.9\linewidth]{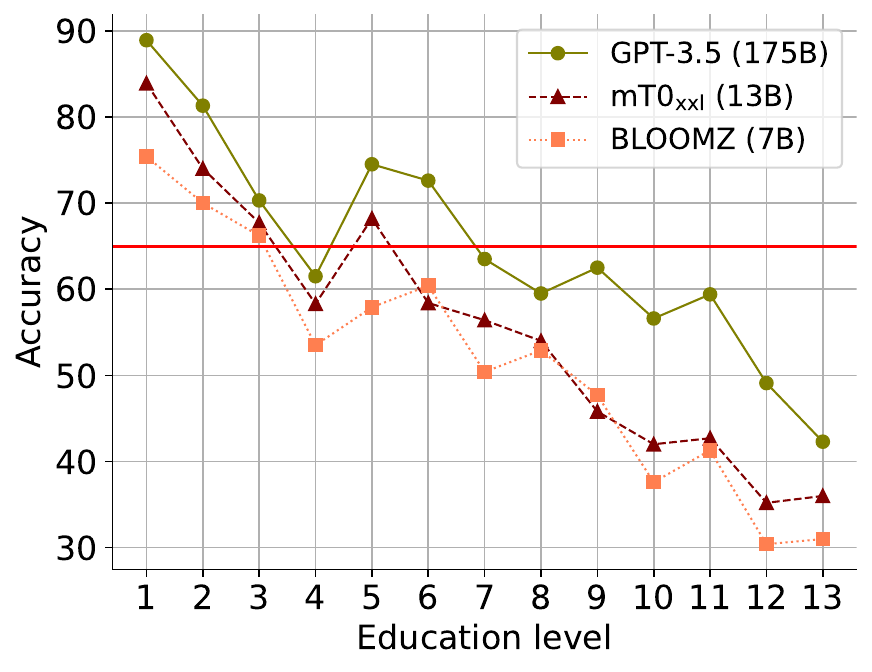} 
    \caption{Fine-grained accuracy (\%) of GPT-3.5, mT0$_\text{xxl}$, and BLOOMZ in the Indonesian language subject area. The horizontal line depicts the passing score of 65, and the education level of 13 refers to the university entrance exam.}
    \label{fig:indo}
\end{figure}

\paragraph{Indonesian language proficiency of LLMs} As discussed in Section~\ref{sec:data_construct}, \texttt{IndoMMLU} specifically includes Indonesian language exams for all grades and education levels, allowing us to assess the Indonesian language proficiency of LLMs.  Figure~\ref{fig:indo} illustrates that GPT-3.5 achieves its highest accuracy in grade 1, approaching 90\%. However, as the education level increases, the model's performance gradually declines. For grades 3 and above,
the scores fall below 75, and for classes 7 and above, GPT-3.5 fails to pass the exams. We observe that this trend is similar for mT0$_\text{xxl}$ and BLOOMZ, which only pass grades 1, 2, and 3. This fine-grained evaluation provides a valuable benchmark for LLM proficiency in Indonesian.

\paragraph{LLM performance on local languages and cultures}

\begin{figure}[t]
    \centering
    \includegraphics[width=\linewidth]{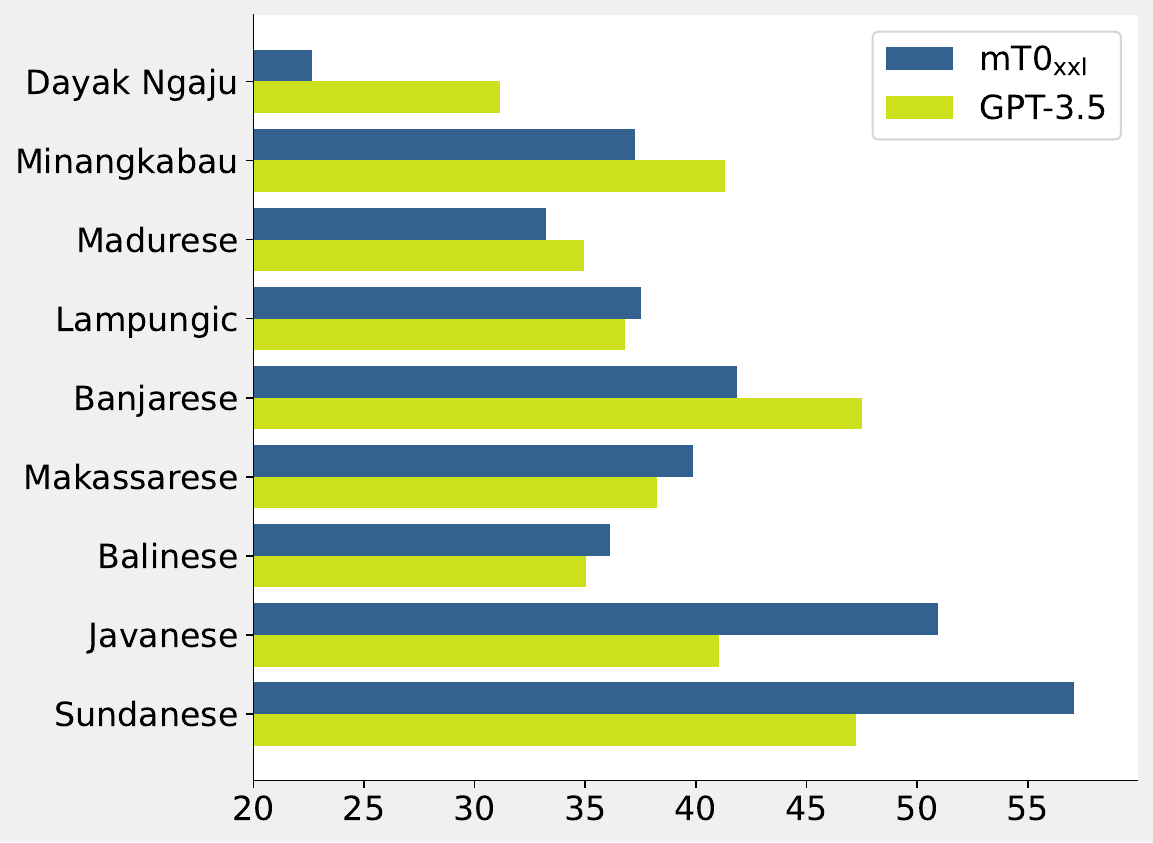} 
    \caption{Zero-shot performance of mT0$_\text{xxl}$ and GPT-3.5 in local languages and cultures subjects.}
    \label{fig:performance_local}
\end{figure}

It is interesting to observe in Table~\ref{tab:result} that despite having only 13B parameters, mT0$_\text{xxl}$ achieves the highest accuracy on local languages and cultures. On the other hand, GPT-3.5 with 175B parameters achieves competitive accuracy, just 0.3 absolute points lower than mT0$_\text{xxl}$. To further investigate this, Figure~\ref{fig:performance_local} displays the accuracy scores of each local language and culture subject, revealing that both mT0$_\text{xxl}$ and GPT-3.5 excel in different subject areas. mT0$_\text{xxl}$ shows greater familiarity with Javanese and Sundanese, with a disparity of $+$10 for both subjects compared to GPT-3.5. GPT-3.5 performs better in Dayak Ngaju, Banjarese, and Minangkabau culture.

\subsection{Analysis}

\begin{figure}[t]
    \centering
    \includegraphics[width=0.9\linewidth]{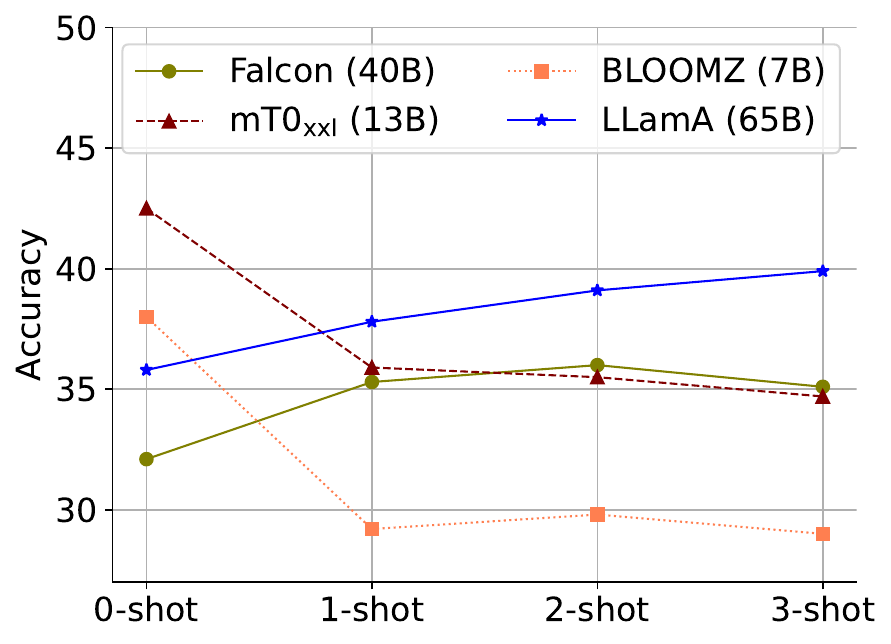} 
    \caption{Few-shot performance (\% accuracy) of mT0$_\text{xxl}$, BLOOMZ, Falcon, and LLaMA, averaged across all subject areas.}
    \label{fig:fewshot}
\end{figure}

\paragraph{Few shot performance} Providing several questions and the answer key in prompts has been reported to improve model performance \cite{karimi-mahabadi-etal-2022-prompt,hendrycksmeasuring}. We run similar experiments with our top-4 best open-source models and observe mixed outcomes in Figure~\ref{fig:fewshot}.\footnote{Refer to the Appendix for details of the prompts.} Few-shot inference does not yield improvements in instruction-tuned models like mT0 and BLOOMZ, as evidenced by a decrease in accuracy. In contrast, the pure LLMs Falcon and LLaMA show better performance with few-shot inference compared to zero-shot. These findings align with those of \citet{liu2023m3ke,li2023cmmlu}, where few-shot prompts may lead to unnatural inferences for instruction-tuned models.

\paragraph{Model confidence}

Given the top three models in Table~\ref{tab:result}, we assess whether their confidence predictions (i.e.~the predicted likelihood of the predicted answer being correct) corresponds to the actual accuracy across 64 tasks. This uncertainty calibration gives us hints about the model's reliability and how to use them appropriately in real-world settings. For mT0 and BLOOMZ, the confidence score is determined through softmax normalization over probabilities of the multiple-choice options. For GPT-3.5, we adopt the approach described by \citet{si2022prompting, wang2022self}, using a high-temperature value (0.7) during decoding.  For each question, we generate $n$ different outputs and measure self-consistency. The probability of a multiple-choice option is calculated based on the output frequency. In this experiment, we use $n=7$, and choose the most frequently-occurring answer as the final prediction.

\begin{figure}[t]
    \centering
    \includegraphics[width=0.9\linewidth]{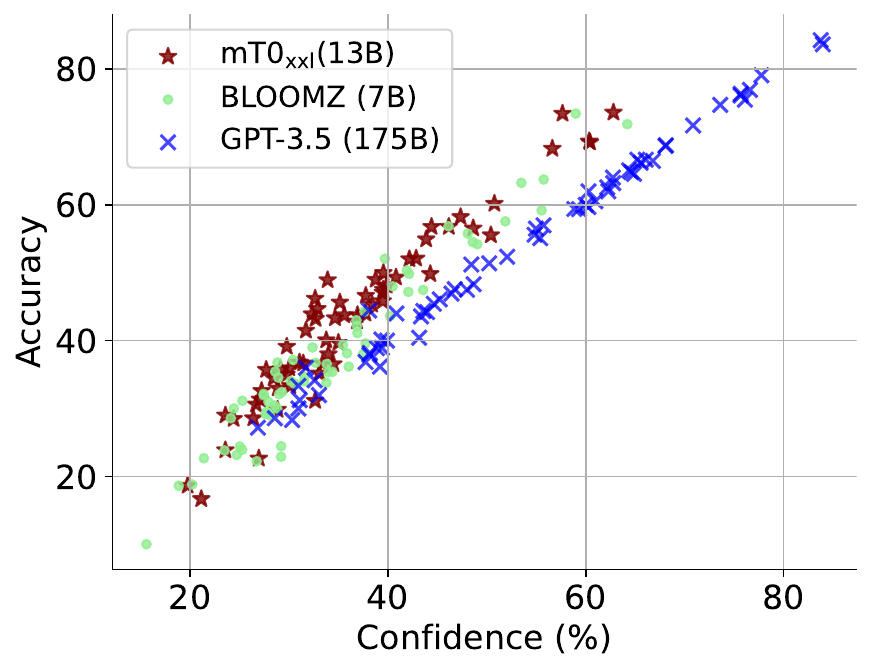} 
    \caption{Zero-shot calibration of mT0$_\text{xxl}$, BLOOMZ, and GPT-3.5 across 64 tasks. The average standard deviations of the confidence scores across all data points are 36.5, 26.4, and 43.9, respectively}
    \label{fig:conf1}
     \vspace{-0cm}
\end{figure}

We average the confidence scores across the 64 tasks, and display the calibration of mT0, BLOOMZ, and GPT-3.5 in Figure~\ref{fig:conf1}. We observe that all three models are well-calibrated, with correlation scores of $r > 0.85$.

\begin{figure}[t]
    \centering
    \includegraphics[width=\linewidth]{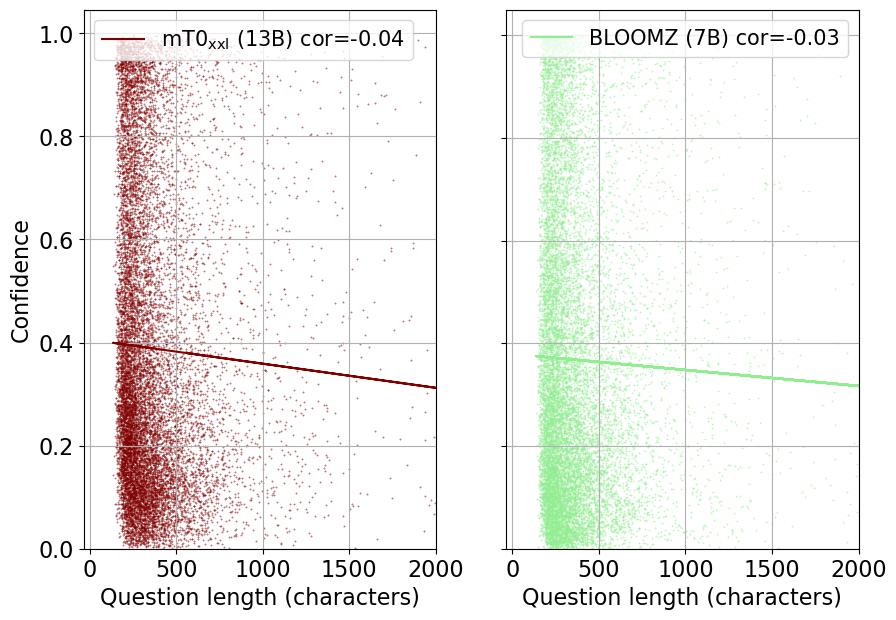} 
    \caption{Correlation between question difficulty and question length.}
    \label{fig:conf2}
\end{figure}

Additionally, we examine the relationship between confidence scores and question length, as depicted in Figure~\ref{fig:conf2}. We found a very weak correlation for both mT0 and BLOOMZ. It is worth noting that the confidence score can also be interpreted as a measure of question difficulty, based on which question length appears to have no bearing on difficulty.

\paragraph{Impact of negation}

In Indonesian school exam questions, the use of negation is common to enhance question difficulty and assess students' reasoning abilities. Similarly, in the field of NLP, negation is known to increase the difficulty of NLP tasks \cite{truong-etal-2022-improving}. To investigate the impact of negation, we employ a simple string-matching strategy to identify questions that contain negations within each subject area.\footnote{To identify negation, we use strings \textit{kecuali} \gl{except}, \textit{yang bukan} \gl{which is not}, and \textit{yang tidak} \gl{which is not}.} We then break down the accuracy for the top three models (GPT-3.5, mT0, and BLOOMZ) based on the presence or absence of negation. Among the subject areas, Indonesian language and social science are the most prevalent in employing negation, accounting for approximately 10\% in each group. Through manual observation of 100 random samples, we verified that 85\% of these questions indeed contained negation. 

Table~\ref{tab:negation} shows the effects of negation on \texttt{IndoMMLU} accuracy.
For the Indonesian language subject area, negated questions prove to be more challenging, with a decrease in accuracy ranging from $-$4 to $-$10. In social science, mT0 and BLOOMZ are similarly more accurate over questions without negation. Compared to mT0, however, BLOOMZ is less robust to negation, as indicated by the $-$5 accuracy drop.


\begin{table}[t]
    \centering
    \resizebox{\linewidth}{!}{
        \begin{tabular}{lcc}
        \toprule
            \textbf{Model} & \textbf{W/ negation} & \textbf{W/o negation} \\
            \midrule
            \multicolumn{3}{l}{\textit{Indonesian language}} \\ 
            GPT-3.5 (175B) & 58.0 & \textbf{62.7} \\
            mT0$_\text{xxl}$ (13B) & 47.9 & \textbf{53.1} \\
            BLOOMZ (7B) & 39.3 & \textbf{50.1} \\
            \midrule
            \multicolumn{3}{l}{\textit{Social science}}  \\
            GPT-3.5 (175B) & \textbf{66.2} & 63.0 \\
            mT0$_\text{xxl}$ (13B)  & \textbf{48.2} & 47.1 \\
            BLOOMZ (7B)  & 43.3 & \textbf{48.2} \\

        \bottomrule
        \end{tabular}
    }
    \caption{Accuracy (\%) for questions with and without negation in the Indonesian language and social science subject areas.}
    \label{tab:negation}
\end{table}

\section{Discussion}

If LLMs are to be deployed in diverse contexts, it is critical to have more work on evaluation for different languages and cultures. In Table~\ref{tab:stat} we observed that the models struggle to answer questions that pertain to local languages and cultures across all levels of education in Indonesia. Minangkabau culture in particular is taught and assessed in the Indonesian language, and yet the limited performance in answering questions relating to it underscores a lack of cultural knowledge, despite reasonable results for the Indonesian language.


We also argue that education science should play a more central role in the future evaluation of LLMs. Current NLP work has mostly focused on developing larger models with different techniques and architectures, and evaluation has primarily been in terms of specific NLP tasks. Education science has decades of experience in designing assessments to evaluate student progress through painstakingly-designed comprehensive tests, which the NLP community should better engage with. With \texttt{IndoMMLU}, we have shown that exam questions across fine-grained educational levels offer a more profound comprehension of model proficiency in the Indonesian language, while also revealing potential areas for improvement.



\section{Conclusion}

In this paper, we presented \texttt{IndoMMLU}, a multi-task language understanding benchmark for real-world evaluation of knowledge in the Indonesian context. By leveraging education level metadata, we found that current LLMs like GPT-3.5 are only able to pass primary school exams in Indonesia, while smaller models struggle across nearly in all education levels. Notably, none of the 24 evaluated models perform well in the domain of local languages and cultures, highlighting the need for further research in this direction.

\section*{Limitations}

Despite being the largest question-answering dataset in the Indonesian context,  \texttt{IndoMMLU} still has some limitations, in that it lacks: (1) multimodal questions; (2) arithmetic reasoning tasks; and (3) essay-style questions. First, \texttt{IndoMMLU} is comprised  solely of text-based questions, and questions with tables and figures are discarded to simplify data collection. We specifically exclude math questions as they are already well covered by existing English math reasoning benchmarks. We suggest that essay questions enable a deeper assessment of comprehension and critical thinking, but that methods for evaluating essay quality across education levels in languages other than English are severely lacking.

\section*{Ethical Considerations}

The \texttt{IndoMMLU} dataset used in our study is collected from publicly-available web resources. In compliance with the Indonesian Copyright Law number 28 year 2014, specifically article 44, the use, retrieval, reproduction, and/or modification of works and/or related rights products, in whole or in substantial part, is not considered a copyright infringement if the source is fully cited or mentioned for educational and research purposes.\footnote{\url{https://wipolex-res.wipo.int/edocs/lexdocs/laws/en/id/id064en.pdf}}

Regarding our experimental results, it is important to note that they do not provide a definitive answer as to the relative abilities of LLMs, and we caution readers against overinterpreting the findings. While we conclude that GPT-3.5 demonstrates proficiency in passing primary school exams in Indonesia based on \texttt{IndoMMLU}, it is essential to consider potential contamination in GPT-3.5's pre-training data, which could impact the results. Furthermore, it is worth noting that real-world student assessments encompass not only multiple-choice questions but also practical exams, laboratory work, and essay writing.


\bibliography{custom,anthology}

\begin{thebibliography}{58}
\expandafter\ifx\csname natexlab\endcsname\relax\def\natexlab#1{#1}\fi

\bibitem[{Baradaran et~al.(2022)Baradaran, Ghiasi, and
  Amirkhani}]{baradaran2022survey}
Razieh Baradaran, Razieh Ghiasi, and Hossein Amirkhani. 2022.
\newblock A survey on machine reading comprehension systems.
\newblock \emph{Natural Language Engineering}, 28(6):683--732.

\bibitem[{Bisk et~al.(2020)Bisk, Zellers, Gao, Choi et~al.}]{bisk2020piqa}
Yonatan Bisk, Rowan Zellers, Jianfeng Gao, Yejin Choi, et~al. 2020.
\newblock {PIQA}: Reasoning about physical commonsense in natural language.
\newblock In \emph{Proceedings of the AAAI conference on artificial
  intelligence}, pages 7432--7439.

\bibitem[{Cahyawijaya et~al.(2023)Cahyawijaya, Lovenia, Aji, Winata, Wilie,
  Koto, Mahendra, Wibisono, Romadhony, Vincentio, Santoso, Moeljadi, Wirawan,
  Hudi, Wicaksono, Parmonangan, Alfina, Putra, Rahmadani, Oenang, Septiandri,
  Jaya, Dhole, Suryani, Putri, Su, Stevens, Nityasya, Adilazuarda, Hadiwijaya,
  Diandaru, Yu, Ghifari, Dai, Xu, Damapuspita, Wibowo, Tho, Karo, Fatyanosa,
  Ji, Neubig, Baldwin, Ruder, Fung, Sujaini, Sakti, , and
  Purwarianti}]{cahyawijaya2022nusacrowd}
Samuel Cahyawijaya, Holy Lovenia, Alham~Fikri Aji, Genta~Indra Winata, Bryan
  Wilie, Fajri Koto, Rahmad Mahendra, Christian Wibisono, Ade Romadhony,
  Karissa Vincentio, Jennifer Santoso, David Moeljadi, Cahya Wirawan,
  Frederikus Hudi, Muhammad~Satrio Wicaksono, Ivan~Halim Parmonangan, Ika
  Alfina, Ilham~Firdausi Putra, Samsul Rahmadani, Yulianti Oenang, Ali~Akbar
  Septiandri, James Jaya, Kaustubh Dhole, Arie Suryani, Rifki~Afina Putri, Dan
  Su, Keith~David Stevens, Made~Nindyatama Nityasya, Muhammad~Farid
  Adilazuarda, Ryan~Ignatius Hadiwijaya, Ryandito Diandaru, Tiezheng Yu, Vito
  Ghifari, Wenliang Dai, Yan Xu, Dyah~Inastra Damapuspita, Haryo~Akbarianto
  Wibowo, Cuk Tho, Ichwanul Muslim~Karo Karo, Tirana~Noor Fatyanosa, Ziwei Ji,
  Graham Neubig, Timothy Baldwin, Sebastian Ruder, Pascale Fung, Herry Sujaini,
  Sakriani Sakti, , and Ayu Purwarianti. 2023.
\newblock {NusaCrowd}: Open source initiative for {Indonesian NLP} resources.
\newblock In \emph{Findings of the Association for Computational Linguistics:
  ACL 2023}.

\bibitem[{Cahyawijaya et~al.(2021)Cahyawijaya, Winata, Wilie, Vincentio, Li,
  Kuncoro, Ruder, Lim, Bahar, Khodra, Purwarianti, and
  Fung}]{cahyawijaya-etal-2021-indonlg}
Samuel Cahyawijaya, Genta~Indra Winata, Bryan Wilie, Karissa Vincentio,
  Xiaohong Li, Adhiguna Kuncoro, Sebastian Ruder, Zhi~Yuan Lim, Syafri Bahar,
  Masayu Khodra, Ayu Purwarianti, and Pascale Fung. 2021.
\newblock \href {https://doi.org/10.18653/v1/2021.emnlp-main.699} {{I}ndo{NLG}:
  Benchmark and resources for evaluating {I}ndonesian natural language
  generation}.
\newblock In \emph{Proceedings of the 2021 Conference on Empirical Methods in
  Natural Language Processing}, pages 8875--8898, Online and Punta Cana,
  Dominican Republic. Association for Computational Linguistics.

\bibitem[{Chalkidis(2023)}]{chalkidis2023chatgpt}
Ilias Chalkidis. 2023.
\newblock {ChatGPT} may pass the bar exam soon, but has a long way to go for
  the {LexGLUE} benchmark.
\newblock \emph{arXiv preprint arXiv:2304.12202}.

\bibitem[{Choi et~al.(2023)Choi, Hickman, Monahan, and
  Schwarcz}]{choi2023chatgpt}
Jonathan~H Choi, Kristin~E Hickman, Amy Monahan, and Daniel Schwarcz. 2023.
\newblock {ChatGPT} goes to law school.
\newblock \emph{Available at SSRN}.

\bibitem[{Clark et~al.(2020)Clark, Choi, Collins, Garrette, Kwiatkowski,
  Nikolaev, and Palomaki}]{clark-etal-2020-tydi}
Jonathan~H. Clark, Eunsol Choi, Michael Collins, Dan Garrette, Tom Kwiatkowski,
  Vitaly Nikolaev, and Jennimaria Palomaki. 2020.
\newblock \href {https://doi.org/10.1162/tacl_a_00317} {{T}y{D}i {QA}: A
  benchmark for information-seeking question answering in typologically diverse
  languages}.
\newblock \emph{Transactions of the Association for Computational Linguistics},
  8:454--470.

\bibitem[{Clark et~al.(2018)Clark, Cowhey, Etzioni, Khot, Sabharwal, Schoenick,
  and Tafjord}]{clark2018think}
Peter Clark, Isaac Cowhey, Oren Etzioni, Tushar Khot, Ashish Sabharwal, Carissa
  Schoenick, and Oyvind Tafjord. 2018.
\newblock Think you have solved question answering? {Try} {ARC}, the {AI2}
  reasoning challenge.
\newblock \emph{arXiv preprint arXiv:1803.05457}.

\bibitem[{Cobbe et~al.(2021)Cobbe, Kosaraju, Bavarian, Chen, Jun, Kaiser,
  Plappert, Tworek, Hilton, Nakano et~al.}]{cobbe2021training}
Karl Cobbe, Vineet Kosaraju, Mohammad Bavarian, Mark Chen, Heewoo Jun, Lukasz
  Kaiser, Matthias Plappert, Jerry Tworek, Jacob Hilton, Reiichiro Nakano,
  et~al. 2021.
\newblock Training verifiers to solve math word problems.
\newblock \emph{arXiv preprint arXiv:2110.14168}.

\bibitem[{Conneau et~al.(2020)Conneau, Khandelwal, Goyal, Chaudhary, Wenzek,
  Guzm{\'a}n, Grave, Ott, Zettlemoyer, and
  Stoyanov}]{conneau-etal-2020-unsupervised}
Alexis Conneau, Kartikay Khandelwal, Naman Goyal, Vishrav Chaudhary, Guillaume
  Wenzek, Francisco Guzm{\'a}n, Edouard Grave, Myle Ott, Luke Zettlemoyer, and
  Veselin Stoyanov. 2020.
\newblock \href {https://doi.org/10.18653/v1/2020.acl-main.747} {Unsupervised
  cross-lingual representation learning at scale}.
\newblock In \emph{Proceedings of the 58th Annual Meeting of the Association
  for Computational Linguistics}, pages 8440--8451, Online. Association for
  Computational Linguistics.

\bibitem[{Devlin et~al.(2019)Devlin, Chang, Lee, and
  Toutanova}]{devlin-etal-2019-bert}
Jacob Devlin, Ming-Wei Chang, Kenton Lee, and Kristina Toutanova. 2019.
\newblock \href {https://doi.org/10.18653/v1/N19-1423} {{BERT}: Pre-training of
  deep bidirectional transformers for language understanding}.
\newblock In \emph{Proceedings of the 2019 Conference of the North {A}merican
  Chapter of the Association for Computational Linguistics: Human Language
  Technologies, Volume 1 (Long and Short Papers)}, pages 4171--4186,
  Minneapolis, Minnesota. Association for Computational Linguistics.

\bibitem[{Gehrmann et~al.(2021)Gehrmann, Adewumi, Aggarwal, Ammanamanchi,
  Aremu, Bosselut, Chandu, Clinciu, Das, Dhole, Du, Durmus, Du{\v{s}}ek,
  Emezue, Gangal, Garbacea, Hashimoto, Hou, Jernite, Jhamtani, Ji, Jolly, Kale,
  Kumar, Ladhak, Madaan, Maddela, Mahajan, Mahamood, Majumder, Martins,
  McMillan-Major, Mille, van Miltenburg, Nadeem, Narayan, Nikolaev,
  Niyongabo~Rubungo, Osei, Parikh, Perez-Beltrachini, Rao, Raunak, Rodriguez,
  Santhanam, Sedoc, Sellam, Shaikh, Shimorina, Sobrevilla~Cabezudo, Strobelt,
  Subramani, Xu, Yang, Yerukola, and Zhou}]{gehrmann-etal-2021-gem}
Sebastian Gehrmann, Tosin Adewumi, Karmanya Aggarwal, Pawan~Sasanka
  Ammanamanchi, Anuoluwapo Aremu, Antoine Bosselut, Khyathi~Raghavi Chandu,
  Miruna-Adriana Clinciu, Dipanjan Das, Kaustubh Dhole, Wanyu Du, Esin Durmus,
  Ond{\v{r}}ej Du{\v{s}}ek, Chris~Chinenye Emezue, Varun Gangal, Cristina
  Garbacea, Tatsunori Hashimoto, Yufang Hou, Yacine Jernite, Harsh Jhamtani,
  Yangfeng Ji, Shailza Jolly, Mihir Kale, Dhruv Kumar, Faisal Ladhak, Aman
  Madaan, Mounica Maddela, Khyati Mahajan, Saad Mahamood, Bodhisattwa~Prasad
  Majumder, Pedro~Henrique Martins, Angelina McMillan-Major, Simon Mille, Emiel
  van Miltenburg, Moin Nadeem, Shashi Narayan, Vitaly Nikolaev, Andre
  Niyongabo~Rubungo, Salomey Osei, Ankur Parikh, Laura Perez-Beltrachini,
  Niranjan~Ramesh Rao, Vikas Raunak, Juan~Diego Rodriguez, Sashank Santhanam,
  Jo{\~a}o Sedoc, Thibault Sellam, Samira Shaikh, Anastasia Shimorina,
  Marco~Antonio Sobrevilla~Cabezudo, Hendrik Strobelt, Nishant Subramani, Wei
  Xu, Diyi Yang, Akhila Yerukola, and Jiawei Zhou. 2021.
\newblock \href {https://doi.org/10.18653/v1/2021.gem-1.10} {The {GEM}
  benchmark: Natural language generation, its evaluation and metrics}.
\newblock In \emph{Proceedings of the 1st Workshop on Natural Language
  Generation, Evaluation, and Metrics (GEM 2021)}, pages 96--120, Online.
  Association for Computational Linguistics.

\bibitem[{Hendrycks et~al.(2021)Hendrycks, Burns, Basart, Zou, Mazeika, Song,
  and Steinhardt}]{hendrycksmeasuring}
Dan Hendrycks, Collin Burns, Steven Basart, Andy Zou, Mantas Mazeika, Dawn
  Song, and Jacob Steinhardt. 2021.
\newblock Measuring massive multitask language understanding.
\newblock In \emph{International Conference on Learning Representations}.

\bibitem[{Hu et~al.(2020)Hu, Ruder, Siddhant, Neubig, Firat, and
  Johnson}]{hu2020xtreme}
Junjie Hu, Sebastian Ruder, Aditya Siddhant, Graham Neubig, Orhan Firat, and
  Melvin Johnson. 2020.
\newblock \href {http://arxiv.org/abs/arXiv:2003.11080v1} {{XTREME}: A
  massively multilingual multi-task benchmark for evaluating cross-lingual
  generalization}.
\newblock In \emph{Proceedings of ICML 2020}.

\bibitem[{Huang et~al.(2019)Huang, Le~Bras, Bhagavatula, and
  Choi}]{huang-etal-2019-cosmos}
Lifu Huang, Ronan Le~Bras, Chandra Bhagavatula, and Yejin Choi. 2019.
\newblock \href {https://doi.org/10.18653/v1/D19-1243} {Cosmos {QA}: Machine
  reading comprehension with contextual commonsense reasoning}.
\newblock In \emph{Proceedings of the 2019 Conference on Empirical Methods in
  Natural Language Processing and the 9th International Joint Conference on
  Natural Language Processing (EMNLP-IJCNLP)}, pages 2391--2401, Hong Kong,
  China. Association for Computational Linguistics.

\bibitem[{Joshi et~al.(2017)Joshi, Choi, Weld, and
  Zettlemoyer}]{joshi-etal-2017-triviaqa}
Mandar Joshi, Eunsol Choi, Daniel Weld, and Luke Zettlemoyer. 2017.
\newblock \href {https://doi.org/10.18653/v1/P17-1147} {{T}rivia{QA}: A large
  scale distantly supervised challenge dataset for reading comprehension}.
\newblock In \emph{Proceedings of the 55th Annual Meeting of the Association
  for Computational Linguistics (Volume 1: Long Papers)}, pages 1601--1611,
  Vancouver, Canada. Association for Computational Linguistics.

\bibitem[{Karimi~Mahabadi et~al.(2022)Karimi~Mahabadi, Zettlemoyer, Henderson,
  Mathias, Saeidi, Stoyanov, and Yazdani}]{karimi-mahabadi-etal-2022-prompt}
Rabeeh Karimi~Mahabadi, Luke Zettlemoyer, James Henderson, Lambert Mathias,
  Marzieh Saeidi, Veselin Stoyanov, and Majid Yazdani. 2022.
\newblock \href {https://doi.org/10.18653/v1/2022.acl-long.254} {Prompt-free
  and efficient few-shot learning with language models}.
\newblock In \emph{Proceedings of the 60th Annual Meeting of the Association
  for Computational Linguistics (Volume 1: Long Papers)}, pages 3638--3652,
  Dublin, Ireland. Association for Computational Linguistics.

\bibitem[{Katz et~al.(2023)Katz, Bommarito, Gao, and Arredondo}]{katz2023gpt}
Daniel~Martin Katz, Michael~James Bommarito, Shang Gao, and Pablo Arredondo.
  2023.
\newblock {GPT-4} passes the bar exam.
\newblock \emph{Available at SSRN 4389233}.

\bibitem[{Koto et~al.(2022{\natexlab{a}})Koto, Baldwin, and
  Lau}]{koto-etal-2022-cloze}
Fajri Koto, Timothy Baldwin, and Jey~Han Lau. 2022{\natexlab{a}}.
\newblock \href {https://doi.org/10.18653/v1/2022.csrr-1.2} {Cloze evaluation
  for deeper understanding of commonsense stories in {I}ndonesian}.
\newblock In \emph{Proceedings of the First Workshop on Commonsense
  Representation and Reasoning (CSRR 2022)}, pages 8--16, Dublin, Ireland.
  Association for Computational Linguistics.

\bibitem[{Koto et~al.(2022{\natexlab{b}})Koto, Baldwin, and
  Lau}]{koto-etal-2022-lipkey}
Fajri Koto, Timothy Baldwin, and Jey~Han Lau. 2022{\natexlab{b}}.
\newblock \href {https://aclanthology.org/2022.coling-1.303} {{L}ip{K}ey: A
  large-scale news dataset for absent keyphrases generation and abstractive
  summarization}.
\newblock In \emph{Proceedings of the 29th International Conference on
  Computational Linguistics}, pages 3427--3437, Gyeongju, Republic of Korea.
  International Committee on Computational Linguistics.

\bibitem[{Koto and Koto(2020)}]{koto-koto-2020-towards}
Fajri Koto and Ikhwan Koto. 2020.
\newblock \href {https://aclanthology.org/2020.paclic-1.17} {Towards
  computational linguistics in {M}inangkabau language: Studies on sentiment
  analysis and machine translation}.
\newblock In \emph{Proceedings of the 34th Pacific Asia Conference on Language,
  Information and Computation}, pages 138--148, Hanoi, Vietnam. Association for
  Computational Linguistics.

\bibitem[{Koto et~al.(2020{\natexlab{a}})Koto, Lau, and
  Baldwin}]{koto-etal-2020-liputan6}
Fajri Koto, Jey~Han Lau, and Timothy Baldwin. 2020{\natexlab{a}}.
\newblock \href {https://aclanthology.org/2020.aacl-main.60} {Liputan6: A
  large-scale {I}ndonesian dataset for text summarization}.
\newblock In \emph{Proceedings of the 1st Conference of the Asia-Pacific
  Chapter of the Association for Computational Linguistics and the 10th
  International Joint Conference on Natural Language Processing}, pages
  598--608, Suzhou, China. Association for Computational Linguistics.

\bibitem[{Koto et~al.(2021)Koto, Lau, and
  Baldwin}]{koto-etal-2021-indobertweet}
Fajri Koto, Jey~Han Lau, and Timothy Baldwin. 2021.
\newblock \href {https://doi.org/10.18653/v1/2021.emnlp-main.833}
  {{I}ndo{BERT}weet: A pretrained language model for {I}ndonesian {T}witter
  with effective domain-specific vocabulary initialization}.
\newblock In \emph{Proceedings of the 2021 Conference on Empirical Methods in
  Natural Language Processing}, pages 10660--10668, Online and Punta Cana,
  Dominican Republic. Association for Computational Linguistics.

\bibitem[{Koto et~al.(2020{\natexlab{b}})Koto, Rahimi, Lau, and
  Baldwin}]{koto-etal-2020-indolem}
Fajri Koto, Afshin Rahimi, Jey~Han Lau, and Timothy Baldwin.
  2020{\natexlab{b}}.
\newblock \href {https://doi.org/10.18653/v1/2020.coling-main.66} {{I}ndo{LEM}
  and {I}ndo{BERT}: A benchmark dataset and pre-trained language model for
  {I}ndonesian {NLP}}.
\newblock In \emph{Proceedings of the 28th International Conference on
  Computational Linguistics}, pages 757--770, Barcelona, Spain (Online).
  International Committee on Computational Linguistics.

\bibitem[{Koto and Rahmaningtyas(2017)}]{koto2017inset}
Fajri Koto and Gemala~Y Rahmaningtyas. 2017.
\newblock Inset lexicon: Evaluation of a word list for {Indonesian} sentiment
  analysis in microblogs.
\newblock In \emph{2017 International Conference on Asian Language Processing
  (IALP)}, pages 391--394. IEEE.

\bibitem[{Kwiatkowski et~al.(2019)Kwiatkowski, Palomaki, Redfield, Collins,
  Parikh, Alberti, Epstein, Polosukhin, Devlin, Lee, Toutanova, Jones, Kelcey,
  Chang, Dai, Uszkoreit, Le, and Petrov}]{kwiatkowski-etal-2019-natural}
Tom Kwiatkowski, Jennimaria Palomaki, Olivia Redfield, Michael Collins, Ankur
  Parikh, Chris Alberti, Danielle Epstein, Illia Polosukhin, Jacob Devlin,
  Kenton Lee, Kristina Toutanova, Llion Jones, Matthew Kelcey, Ming-Wei Chang,
  Andrew~M. Dai, Jakob Uszkoreit, Quoc Le, and Slav Petrov. 2019.
\newblock \href {https://doi.org/10.1162/tacl_a_00276} {Natural questions: A
  benchmark for question answering research}.
\newblock \emph{Transactions of the Association for Computational Linguistics},
  7:452--466.

\bibitem[{Lai et~al.(2017)Lai, Xie, Liu, Yang, and Hovy}]{lai-etal-2017-race}
Guokun Lai, Qizhe Xie, Hanxiao Liu, Yiming Yang, and Eduard Hovy. 2017.
\newblock \href {https://doi.org/10.18653/v1/D17-1082} {{RACE}: Large-scale
  {R}e{A}ding comprehension dataset from examinations}.
\newblock In \emph{Proceedings of the 2017 Conference on Empirical Methods in
  Natural Language Processing}, pages 785--794, Copenhagen, Denmark.
  Association for Computational Linguistics.

\bibitem[{Li et~al.(2023{\natexlab{a}})Li, Koto, Wu, Aji, and
  Baldwin}]{li2023bactrian}
Haonan Li, Fajri Koto, Minghao Wu, Alham~Fikri Aji, and Timothy Baldwin.
  2023{\natexlab{a}}.
\newblock Bactrian-{X}: A multilingual replicable instruction-following model
  with low-rank adaptation.
\newblock \emph{arXiv preprint arXiv:2305.15011}.

\bibitem[{Li et~al.(2023{\natexlab{b}})Li, Zhang, Koto, Yang, Zhao, Gong, Duan,
  and Baldwin}]{li2023cmmlu}
Haonan Li, Yixuan Zhang, Fajri Koto, Yifei Yang, Hai Zhao, Yeyun Gong, Nan
  Duan, and Timothy Baldwin. 2023{\natexlab{b}}.
\newblock {CMMLU}: Measuring massive multitask language understanding in
  chinese.
\newblock \emph{arXiv preprint arXiv:2306.09212}.

\bibitem[{Liang et~al.(2020)Liang, Duan, Gong, Wu, Guo, Qi, Gong, Shou, Jiang,
  Cao, Fan, Zhang, Agrawal, Cui, Wei, Bharti, Qiao, Chen, Wu, Liu, Yang,
  Campos, Majumder, and Zhou}]{liang-etal-2020-xglue}
Yaobo Liang, Nan Duan, Yeyun Gong, Ning Wu, Fenfei Guo, Weizhen Qi, Ming Gong,
  Linjun Shou, Daxin Jiang, Guihong Cao, Xiaodong Fan, Ruofei Zhang, Rahul
  Agrawal, Edward Cui, Sining Wei, Taroon Bharti, Ying Qiao, Jiun-Hung Chen,
  Winnie Wu, Shuguang Liu, Fan Yang, Daniel Campos, Rangan Majumder, and Ming
  Zhou. 2020.
\newblock \href {https://doi.org/10.18653/v1/2020.emnlp-main.484} {{XGLUE}: A
  new benchmark dataset for cross-lingual pre-training, understanding and
  generation}.
\newblock In \emph{Proceedings of the 2020 Conference on Empirical Methods in
  Natural Language Processing (EMNLP)}, pages 6008--6018, Online. Association
  for Computational Linguistics.

\bibitem[{Lin et~al.(2021)Lin, Mihaylov, Artetxe, Wang, Chen, Simig, Ott,
  Goyal, Bhosale, Du et~al.}]{lin2021few}
Xi~Victoria Lin, Todor Mihaylov, Mikel Artetxe, Tianlu Wang, Shuohui Chen,
  Daniel Simig, Myle Ott, Naman Goyal, Shruti Bhosale, Jingfei Du, et~al. 2021.
\newblock Few-shot learning with multilingual language models.
\newblock \emph{arXiv preprint arXiv:2112.10668}.

\bibitem[{Liu et~al.(2023{\natexlab{a}})Liu, Koto, Baldwin, and
  Gurevych}]{liu2023multilingual}
Chen~Cecilia Liu, Fajri Koto, Timothy Baldwin, and Iryna Gurevych.
  2023{\natexlab{a}}.
\newblock Are multilingual llms culturally-diverse reasoners? an investigation
  into multicultural proverbs and sayings.
\newblock \emph{arXiv preprint arXiv:2309.08591}.

\bibitem[{Liu et~al.(2023{\natexlab{b}})Liu, Jin, Ren, Yu, Dong, Peng, Zhang,
  Peng, Zhang, Lyu et~al.}]{liu2023m3ke}
Chuang Liu, Renren Jin, Yuqi Ren, Linhao Yu, Tianyu Dong, Xiaohan Peng, Shuting
  Zhang, Jianxiang Peng, Peiyi Zhang, Qingqing Lyu, et~al. 2023{\natexlab{b}}.
\newblock {M3KE}: A massive multi-level multi-subject knowledge evaluation
  benchmark for chinese large language models.
\newblock \emph{arXiv preprint arXiv:2305.10263}.

\bibitem[{Mishra et~al.(2022)Mishra, Mitra, Varshney, Sachdeva, Clark, Baral,
  and Kalyan}]{mishra-etal-2022-numglue}
Swaroop Mishra, Arindam Mitra, Neeraj Varshney, Bhavdeep Sachdeva, Peter Clark,
  Chitta Baral, and Ashwin Kalyan. 2022.
\newblock \href {https://doi.org/10.18653/v1/2022.acl-long.246} {{N}um{GLUE}: A
  suite of fundamental yet challenging mathematical reasoning tasks}.
\newblock In \emph{Proceedings of the 60th Annual Meeting of the Association
  for Computational Linguistics (Volume 1: Long Papers)}, pages 3505--3523,
  Dublin, Ireland. Association for Computational Linguistics.

\bibitem[{Muennighoff et~al.(2022)Muennighoff, Wang, Sutawika, Roberts,
  Biderman, Scao, Bari, Shen, Yong, Schoelkopf
  et~al.}]{muennighoff2022crosslingual}
Niklas Muennighoff, Thomas Wang, Lintang Sutawika, Adam Roberts, Stella
  Biderman, Teven~Le Scao, M~Saiful Bari, Sheng Shen, Zheng-Xin Yong, Hailey
  Schoelkopf, et~al. 2022.
\newblock Crosslingual generalization through multitask finetuning.
\newblock \emph{arXiv preprint arXiv:2211.01786}.

\bibitem[{Novak(1988)}]{novak1988learning}
Joseph Novak. 1988.
\newblock Learning science and the science of learning.
\newblock \emph{Studies in Science Education}, 15(1):77--101.

\bibitem[{OpenAI(2023)}]{OpenAI2023GPT4TR}
OpenAI. 2023.
\newblock {GPT-4} technical report.
\newblock \emph{ArXiv}, abs/2303.08774.

\bibitem[{Ouyang et~al.(2022)Ouyang, Wu, Jiang, Almeida, Wainwright, Mishkin,
  Zhang, Agarwal, Slama, Ray et~al.}]{ouyang2022training}
Long Ouyang, Jeffrey Wu, Xu~Jiang, Diogo Almeida, Carroll Wainwright, Pamela
  Mishkin, Chong Zhang, Sandhini Agarwal, Katarina Slama, Alex Ray, et~al.
  2022.
\newblock Training language models to follow instructions with human feedback.
\newblock \emph{Advances in Neural Information Processing Systems},
  35:27730--27744.

\bibitem[{Penedo et~al.(2023)Penedo, Malartic, Hesslow, Cojocaru, Cappelli,
  Alobeidli, Pannier, Almazrouei, and Launay}]{penedo2023refinedweb}
Guilherme Penedo, Quentin Malartic, Daniel Hesslow, Ruxandra Cojocaru,
  Alessandro Cappelli, Hamza Alobeidli, Baptiste Pannier, Ebtesam Almazrouei,
  and Julien Launay. 2023.
\newblock The {RefinedWeb} dataset for {Falcon LLM}: Outperforming curated
  corpora with web data, and web data only.
\newblock \emph{arXiv preprint arXiv:2306.01116}.

\bibitem[{Purwarianti and Crisdayanti(2019)}]{purwarianti2019improving}
Ayu Purwarianti and Ida Ayu Putu~Ari Crisdayanti. 2019.
\newblock Improving bi-{LSTM} performance for {Indonesian} sentiment analysis
  using paragraph vector.
\newblock In \emph{2019 International Conference of Advanced Informatics:
  Concepts, Theory and Applications (ICAICTA)}, pages 1--5. IEEE.

\bibitem[{Purwarianti et~al.(2007)Purwarianti, Tsuchiya, and
  Nakagawa}]{purwarianti2007machine}
Ayu Purwarianti, Masatoshi Tsuchiya, and Seiichi Nakagawa. 2007.
\newblock A machine learning approach for {Indonesian} question answering
  system.
\newblock In \emph{Artificial Intelligence and Applications}, pages 573--578.

\bibitem[{Putri and Oh(2022)}]{putri-oh-2022-idk}
Rifki~Afina Putri and Alice Oh. 2022.
\newblock \href {https://aclanthology.org/2022.emnlp-main.465} {{IDK}-{MRC}:
  Unanswerable questions for {I}ndonesian machine reading comprehension}.
\newblock In \emph{Proceedings of the 2022 Conference on Empirical Methods in
  Natural Language Processing}, pages 6918--6933, Abu Dhabi, United Arab
  Emirates. Association for Computational Linguistics.

\bibitem[{Ruder et~al.(2021)Ruder, Constant, Botha, Siddhant, Firat, Fu, Liu,
  Hu, Garrette, Neubig, and Johnson}]{ruder-etal-2021-xtreme}
Sebastian Ruder, Noah Constant, Jan Botha, Aditya Siddhant, Orhan Firat, Jinlan
  Fu, Pengfei Liu, Junjie Hu, Dan Garrette, Graham Neubig, and Melvin Johnson.
  2021.
\newblock \href {https://doi.org/10.18653/v1/2021.emnlp-main.802}
  {{XTREME}-{R}: Towards more challenging and nuanced multilingual evaluation}.
\newblock In \emph{Proceedings of the 2021 Conference on Empirical Methods in
  Natural Language Processing}, pages 10215--10245, Online and Punta Cana,
  Dominican Republic. Association for Computational Linguistics.

\bibitem[{Ryznar(2023)}]{ryznar2023exams}
Margaret Ryznar. 2023.
\newblock Exams in the time of {ChatGPT}.
\newblock \emph{Washington and Lee Law Review Online}, 80(5):305.

\bibitem[{Sakaguchi et~al.(2021)Sakaguchi, Bras, Bhagavatula, and
  Choi}]{sakaguchi2021winogrande}
Keisuke Sakaguchi, Ronan~Le Bras, Chandra Bhagavatula, and Yejin Choi. 2021.
\newblock Winogrande: An adversarial {Winograd} schema challenge at scale.
\newblock \emph{Communications of the ACM}, 64(9):99--106.

\bibitem[{Saputri et~al.(2018)Saputri, Mahendra, and
  Adriani}]{saputri2018emotion}
Mei~Silviana Saputri, Rahmad Mahendra, and Mirna Adriani. 2018.
\newblock Emotion classification on {Indonesian} {Twitter} dataset.
\newblock In \emph{2018 International Conference on Asian Language Processing
  (IALP)}, pages 90--95. IEEE.

\bibitem[{Sengupta et~al.(2023)Sengupta, Sahu, Jia, Katipomu, Li, Koto, Afzal,
  Kamboj, Pandit, Pal et~al.}]{sengupta2023jais}
Neha Sengupta, Sunil~Kumar Sahu, Bokang Jia, Satheesh Katipomu, Haonan Li,
  Fajri Koto, Osama~Mohammed Afzal, Samta Kamboj, Onkar Pandit, Rahul Pal,
  et~al. 2023.
\newblock Jais and jais-chat: Arabic-centric foundation and instruction-tuned
  open generative large language models.
\newblock \emph{arXiv preprint arXiv:2308.16149}.

\bibitem[{Si et~al.(2022)Si, Gan, Yang, Wang, Wang, Boyd-Graber, and
  Wang}]{si2022prompting}
Chenglei Si, Zhe Gan, Zhengyuan Yang, Shuohang Wang, Jianfeng Wang, Jordan
  Boyd-Graber, and Lijuan Wang. 2022.
\newblock Prompting {GPT-3} to be reliable.
\newblock \emph{arXiv preprint arXiv:2210.09150}.

\bibitem[{Talmor et~al.(2019)Talmor, Herzig, Lourie, and
  Berant}]{talmor-etal-2019-commonsenseqa}
Alon Talmor, Jonathan Herzig, Nicholas Lourie, and Jonathan Berant. 2019.
\newblock \href {https://doi.org/10.18653/v1/N19-1421} {{C}ommonsense{QA}: A
  question answering challenge targeting commonsense knowledge}.
\newblock In \emph{Proceedings of the 2019 Conference of the North {A}merican
  Chapter of the Association for Computational Linguistics: Human Language
  Technologies, Volume 1 (Long and Short Papers)}, pages 4149--4158,
  Minneapolis, Minnesota. Association for Computational Linguistics.

\bibitem[{Touvron et~al.(2023)Touvron, Lavril, Izacard, Martinet, Lachaux,
  Lacroix, Rozi{\`e}re, Goyal, Hambro, Azhar et~al.}]{touvron2023llama}
Hugo Touvron, Thibaut Lavril, Gautier Izacard, Xavier Martinet, Marie-Anne
  Lachaux, Timoth{\'e}e Lacroix, Baptiste Rozi{\`e}re, Naman Goyal, Eric
  Hambro, Faisal Azhar, et~al. 2023.
\newblock {LLaMA}: Open and efficient foundation language models.
\newblock \emph{arXiv preprint arXiv:2302.13971}.

\bibitem[{Truong et~al.(2022)Truong, Baldwin, Cohn, and
  Verspoor}]{truong-etal-2022-improving}
Thinh Truong, Timothy Baldwin, Trevor Cohn, and Karin Verspoor. 2022.
\newblock \href {https://doi.org/10.18653/v1/2022.naacl-main.309} {Improving
  negation detection with negation-focused pre-training}.
\newblock In \emph{Proceedings of the 2022 Conference of the North American
  Chapter of the Association for Computational Linguistics: Human Language
  Technologies}, pages 4188--4193, Seattle, United States. Association for
  Computational Linguistics.

\bibitem[{Wang et~al.(2019)Wang, Pruksachatkun, Nangia, Singh, Michael, Hill,
  Levy, and Bowman}]{wang2019superglue}
Alex Wang, Yada Pruksachatkun, Nikita Nangia, Amanpreet Singh, Julian Michael,
  Felix Hill, Omer Levy, and Samuel Bowman. 2019.
\newblock {SuperGLUE}: A stickier benchmark for general-purpose language
  understanding systems.
\newblock \emph{Advances in neural information processing systems}, 32.

\bibitem[{Wang et~al.(2018)Wang, Singh, Michael, Hill, Levy, and
  Bowman}]{wang-etal-2018-glue}
Alex Wang, Amanpreet Singh, Julian Michael, Felix Hill, Omer Levy, and Samuel
  Bowman. 2018.
\newblock \href {https://doi.org/10.18653/v1/W18-5446} {{GLUE}: A multi-task
  benchmark and analysis platform for natural language understanding}.
\newblock In \emph{Proceedings of the 2018 {EMNLP} Workshop {B}lackbox{NLP}:
  Analyzing and Interpreting Neural Networks for {NLP}}, pages 353--355,
  Brussels, Belgium. Association for Computational Linguistics.

\bibitem[{Wang et~al.(2022)Wang, Wei, Schuurmans, Le, Chi, Narang, Chowdhery,
  and Zhou}]{wang2022self}
Xuezhi Wang, Jason Wei, Dale Schuurmans, Quoc Le, Ed~Chi, Sharan Narang,
  Aakanksha Chowdhery, and Denny Zhou. 2022.
\newblock Self-consistency improves chain of thought reasoning in language
  models.
\newblock \emph{arXiv preprint arXiv:2203.11171}.

\bibitem[{Wilie et~al.(2020)Wilie, Vincentio, Winata, Cahyawijaya, Li, Lim,
  Soleman, Mahendra, Fung, Bahar, and Purwarianti}]{wilie-etal-2020-indonlu}
Bryan Wilie, Karissa Vincentio, Genta~Indra Winata, Samuel Cahyawijaya,
  Xiaohong Li, Zhi~Yuan Lim, Sidik Soleman, Rahmad Mahendra, Pascale Fung,
  Syafri Bahar, and Ayu Purwarianti. 2020.
\newblock \href {https://aclanthology.org/2020.aacl-main.85} {{I}ndo{NLU}:
  Benchmark and resources for evaluating {I}ndonesian natural language
  understanding}.
\newblock In \emph{Proceedings of the 1st Conference of the Asia-Pacific
  Chapter of the Association for Computational Linguistics and the 10th
  International Joint Conference on Natural Language Processing}, pages
  843--857, Suzhou, China. Association for Computational Linguistics.

\bibitem[{Winata et~al.(2023)Winata, Aji, Cahyawijaya, Mahendra, Koto,
  Romadhony, Kurniawan, Moeljadi, Prasojo, Fung, Baldwin, Lau, Sennrich, and
  Ruder}]{winata-etal-2023-nusax}
Genta~Indra Winata, Alham~Fikri Aji, Samuel Cahyawijaya, Rahmad Mahendra, Fajri
  Koto, Ade Romadhony, Kemal Kurniawan, David Moeljadi, Radityo~Eko Prasojo,
  Pascale Fung, Timothy Baldwin, Jey~Han Lau, Rico Sennrich, and Sebastian
  Ruder. 2023.
\newblock \href {https://aclanthology.org/2023.eacl-main.57} {{N}usa{X}:
  Multilingual parallel sentiment dataset for 10 {I}ndonesian local languages}.
\newblock In \emph{Proceedings of the 17th Conference of the European Chapter
  of the Association for Computational Linguistics}, pages 815--834, Dubrovnik,
  Croatia. Association for Computational Linguistics.

\bibitem[{Wolf et~al.(2020)Wolf, Debut, Sanh, Chaumond, Delangue, Moi, Cistac,
  Rault, Louf, Funtowicz, Davison, Shleifer, von Platen, Ma, Jernite, Plu, Xu,
  Le~Scao, Gugger, Drame, Lhoest, and Rush}]{wolf-etal-2020-transformers}
Thomas Wolf, Lysandre Debut, Victor Sanh, Julien Chaumond, Clement Delangue,
  Anthony Moi, Pierric Cistac, Tim Rault, Remi Louf, Morgan Funtowicz, Joe
  Davison, Sam Shleifer, Patrick von Platen, Clara Ma, Yacine Jernite, Julien
  Plu, Canwen Xu, Teven Le~Scao, Sylvain Gugger, Mariama Drame, Quentin Lhoest,
  and Alexander Rush. 2020.
\newblock \href {https://doi.org/10.18653/v1/2020.emnlp-demos.6} {Transformers:
  State-of-the-art natural language processing}.
\newblock In \emph{Proceedings of the 2020 Conference on Empirical Methods in
  Natural Language Processing: System Demonstrations}, pages 38--45, Online.
  Association for Computational Linguistics.

\bibitem[{Zellers et~al.(2019)Zellers, Holtzman, Bisk, Farhadi, and
  Choi}]{zellers-etal-2019-hellaswag}
Rowan Zellers, Ari Holtzman, Yonatan Bisk, Ali Farhadi, and Yejin Choi. 2019.
\newblock \href {https://doi.org/10.18653/v1/P19-1472} {{H}ella{S}wag: Can a
  machine really finish your sentence?}
\newblock In \emph{Proceedings of the 57th Annual Meeting of the Association
  for Computational Linguistics}, pages 4791--4800, Florence, Italy.
  Association for Computational Linguistics.

\end{thebibliography}
\bibliographystyle{acl_natbib}

\clearpage

\appendix

\section{Data Statistics}

Table~\ref{tab:ap1}, Table~\ref{tab:ap2}, and Table~\ref{tab:ap3} provide detailed statistics of the question distribution in \texttt{IndoMMLU}.

\begin{table}[ht]
    \centering
    \resizebox{\linewidth}{!}{
        \begin{tabular}{lrrrrr}
        \toprule
            \textbf{Subjects} & \textbf{SD} & \textbf{SMP} & \textbf{SMA} & \textbf{UE} & \textbf{Total}\\
            \midrule
            Science & 488 & 680 & -- & -- & 1168 \\
            Physics & -- & -- & 297 & -- & 297 \\
            Chemistry & -- & -- & 287 & 398 & 685 \\
            Biology & -- & -- & 457 & 388 & 845 \\
            
            Social science & 300 & 299 & -- & -- & 599 \\
            Geography & -- & -- & 196 & 294 & 490 \\
            Sociology & -- & -- & 295 & 201 & 496 \\
            Economics & -- & -- & 296 & 192 & 488 \\
            History & -- & -- & 300 & 198 & 498 \\
            
            Civics & 99 & 300 & 300 & -- & 699 \\
            Indonesian language & 1125 & 850 & 857 & 381 & 3213 \\
            Balinese & 200 & 123 & 148 & -- & 471 \\
            Makassarese & 98 & 41 & 47 & -- & 186 \\
            Banjarese & 120 & 10 & 14 & -- & 144 \\
            Lampungic & 93 & 30 & 24 & -- & 147 \\
            Madurese & 100 & 93 & 102 & -- & 295 \\
            Sundanese & 718 & 294 & 145 & -- & 1157 \\
            Javanese & 396 & 298 & 298 & -- & 992 \\
            Dayak Ngaju & 109 & -- & -- & -- & 109 \\
            Minangkabau culture & 153 & 46 & -- & -- & 199 \\
            Art & 200 & 200 & 201 & -- & 601 \\
            Sports & 49 & 49 & 50 & -- & 148 \\

            Islam religion & 201 & 202 & 300 & -- & 703 \\
            Christian religion & 50 & 49 & 102 & -- & 201 \\
            Hindu religion & 49 & 52 & 49 & -- & 150 \\
            
            \midrule
            Total & 4548 & 3616 & 4765 & 2052 & 14981 \\  
        \bottomrule
        \end{tabular}
    }
    \caption{Total number of questions for each subject area and education level. ``SD'', ``SMP'', ``SMA'', ``UE'' indicate primary school, junior high school, senior high school, and university entrance tests, respectively.}
    \label{tab:ap1}
\end{table}

\begin{table}[h!]
    \centering
    \resizebox{0.38\linewidth}{!}{
        \begin{tabular}{lr}
        \toprule
            \textbf{Class} & \textbf{\#questions}\\
            \midrule
            1 & 200 \\
            2 & 150 \\
            3 & 195 \\
            4 & 187 \\
            5 & 196 \\
            6 & 197 \\
            7 & 282 \\
            8 & 291 \\
            9 & 277 \\
            10 & 295 \\
            11 & 288 \\
            12 & 274 \\
            12+ & 381 \\
            \midrule
            Total & 3213 \\
        \bottomrule
        \end{tabular}
    }
    \caption{Total number of questions in the Indonesian language subject, including those designated for university entrance tests (12+).}
    \label{tab:ap2}
\end{table}

\begin{table}[h!]
    \centering
    \resizebox{0.9\linewidth}{!}{
        \begin{tabular}{lr}
        \toprule
            \textbf{Category} & \textbf{\#question}\\
            \midrule
                STEM & 2995 \\                
                Social science & 2772  \\
                Humanities & 2301 \\
                Indonesian language & 3213 \\
                Local languages and cultures & 3700 \\
                \midrule
                Total & 14981 \\
        \bottomrule
        \end{tabular}
    }
    \caption{Total number of questions based on subject areas.}
    \label{tab:ap3}
\end{table}

\newpage
\section{Few-shot Prompt}

\begin{figure}[ht!]
    \centering
    \includegraphics[width=\linewidth]{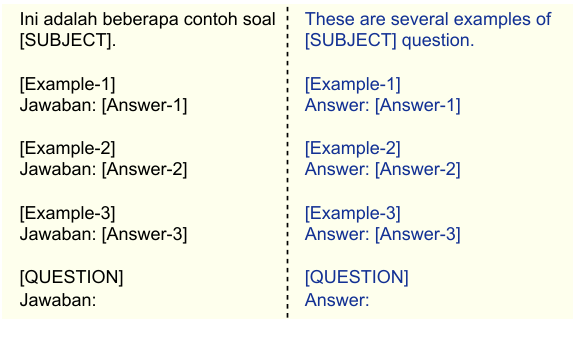} 
    \caption{Illustration of our few-shot prompt template. The English translation on the right is solely for illustrative purposes. In our experiments, we used up to three examples within the prompt. The placeholders \texttt{[SUBJECT]}, \texttt{Example-i}, \texttt{Answer-i}, and \texttt{QUESTION} correspond to the subject, the $i$-th question example, the answer key for the $i$-th question example, and the main question, respectively.}
    \label{fig:ap4}
\end{figure}

\newpage
\onecolumn
\section{Zero-shot Performance Based on the Probability of the Full Generated Answer }
\begin{table*}[ht!]
    \centering
    \resizebox{0.8\linewidth}{!}{
        \begin{tabular}{lC{1cm}C{1.5cm}C{2cm}C{2cm}C{3cm}C{1.5cm}}
        \toprule
       \multirow{2}{*}{\textbf{Model (\#parameters)}}& \multirow{2}{*}{\textbf{STEM}} & \textbf{Social} & \multirow{2}{*}{\textbf{Humanities}} & \textbf{Indonesian} & \textbf{Local languages}  & \multirow{2}{*}{\textbf{Average}} \\
       & & \textbf{Science} & & \textbf{Language} & \textbf{and Cultures} & \\
       \midrule
        Random & 21.9 & 23.4 & 23.5 & 24.4 & 26.6 & 24.4 \\
        \hdashline
        XGLM (564M) & 24.2 & 25.9 & 27.2 & 29.0 & 27.8 & 26.8 \\
        XGLM (1.7B) & 23.7 & 25.4 & 27.1 & 28.4 & 28.9 & 26.9 \\
        XGLM (2.9B) & 23.6 & 25.4 & 28.3 & 28.8 & 28.8 & 26.9 \\
        XGLM (4.5B) & 23.9 & 25.5 & 29.4 & 27.9 & 28.1 & 27.2 \\
        XGLM (7.5B) & 23.5 & 26.0 & 29.4 & 28.6 & 28.9 & 27.6 \\
        \hdashline
        Falcon (7B) & 22.2 & 25.8 & 28.4 & 30.1 & 27.9 & 26.8 \\
        Falcon (40B) & 25.8 & 28.4 & 29.5 & 32.9 & 27.7 & 28.2 \\
        \hdashline
        BLOOMZ (560M) & 23.0 & 24.4 & 23.7 & 27.2 & 26.4 & 24.9 \\
        BLOOMZ (1.1B) & 22.9 & 25.8 & 26.6 & 28.3 & 27.4 & 26.2 \\
        BLOOMZ (1.7B) & 23.7 & 29.8 & 29.7 & 32.8 & 28.1 & 28.3 \\
        BLOOMZ (3B) & 27.6 & 32.5 & 32.6 & 35.0 & 27.4 & 30.0 \\
        BLOOMZ (7.1B) & 26.8 & 32.9 & 33.5 & 36.5 & 28.1 & 30.5 \\
        \hdashline
        mT0$_\text{small}$ (300M)  & 24.0 & 26.1 & 27.0 & 29.8 & 30.8 & 27.8 \\
        mT0$_\text{base}$ (580M)  & 23.9 & 25.5 & 27.6 & 30.1 & 30.5 & 27.7 \\
        mT0$_\text{large}$ (1.2B)  & 25.1 & 27.5 & 27.9 & 33.6 & 29.6 & 28.2 \\
        mT0$_\text{xl}$ (3.7B)  & 28.5 & 36.1 & 35.3 & 40.7 & 34.3 & 34.2 \\
        mT0$_\text{xxl}$ (13B)  & 30.1 & 38.1 & 40.9 & 43.2 & 34.5 & 36.4 \\
        \hdashline
        LLamA (7B) & 23.7 & 25.6 & 28.0 & 29.0 & 28.3 & 27.0 \\
        LLamA (13B) & 24.0 & 25.4 & 27.7 & 29.4 & 29.6 & 27.4 \\
        LLamA (30B) & 24.3 & 26.4 & 29.5 & 29.8 & 28.5 & 27.7 \\
        LLamA (65B) & 26.7 & 29.3 & 32.4 & 32.9 & 29.0 & 29.7 \\
        \hdashline
        Bactrian-X-LLamA (7B) & 23.8 & 25.4 & 28.7 & 29.8 & 28.0 & 27.0 \\
        Bactrian-X-LLamA (13B) & 25.6 & 27.4 & 29.2 & 30.7 & 27.9 & 27.8 \\
        \bottomrule
        \end{tabular}
    }
    \caption{Zero-shot performance (\% accuracy) of large language models based on \textbf{the probability of the full generated answer}, aggregated across education levels. ``Average'' means the average across all subject areas in \texttt{IndoMMLU}. }
    \label{tab:result_full_generated}
\end{table*}

\section{Model Artifacts}
\begin{table}[ht!]
    \centering
    \resizebox{0.5\linewidth}{!}{
        \begin{tabular}{lr}
        \toprule
        \textbf{Models (\#parameters)} & \textbf{Source} \\
        \midrule
        XGLM (564M)  & \texttt{facebook/xglm-564M} \\
        XGLM (1.7B)  & \texttt{facebook/xglm-1.7B}  \\
        XGLM (2.9B)  & \texttt{facebook/xglm-2.9B} \\
        XGLM (4.5B)  & \texttt{facebook/xglm-4.5B} \\
        XGLM (7.5B)  & \texttt{facebook/xglm-7.5B} \\
        \hdashline 
        Falcon (7B)  & \texttt{tiiuae/falcon-7b} \\
        Falcon (40B)  & \texttt{tiiuae/falcon-40b} \\
        \hdashline 
        BLOOMZ (560M)  & \texttt{bigscience/bloomz-560m} \\
        BLOOMZ (1.1B)  & \texttt{bigscience/bloomz-1b1} \\
        BLOOMZ (1.7B)  & \texttt{bigscience/bloomz-1b7} \\
        BLOOMZ (3B)  & \texttt{bigscience/bloomz-3b} \\
        BLOOMZ (7.1B)  & \texttt{bigscience/bloomz-7b1} \\
        \hdashline 
        mT0$_\text{small}$ (300M)   & \texttt{bigscience/mt0-small} \\
        mT0$_\text{base}$ (580M)  & \texttt{bigscience/mt0-base} \\
        mT0$_\text{large}$ (1.2B)   & \texttt{bigscience/mt0-large} \\
        mT0$_\text{xl}$ (3.7B)   & \texttt{bigscience/mt0-xl} \\
        mT0$_\text{xxl}$ (13B)  & \texttt{bigscience/mt0-xxl} \\
        \hdashline 
        LLamA (7B)  & \texttt{decapoda-research/llama-7b-hf} \\
        LLamA (13B)  & \texttt{decapoda-research/llama-13b-hf} \\
        LLamA (30B)  & \texttt{decapoda-research/llama-30b-hf} \\
        LLamA (65B)  & \texttt{huggyllama/llama-65b} \\
        \hdashline 
        Bactrian-X-LLamA (7B)  & \texttt{MBZUAI/bactrian-x-llama-7b-lora} \\
        Bactrian-X-LLamA (13B)  & \texttt{MBZUAI/bactrian-x-llama-13b-lora} \\

        \bottomrule
        \end{tabular}
    }
    \caption{With the exception of GPT--3.5 \cite{ouyang2022training}, all the models used in this study were sourced from Huggingface \cite{wolf-etal-2020-transformers}.}
    \label{tab:models}
\end{table}

\clearpage

\onecolumn
\section{Full Results in Each Subject and Education Level in GPT-3.5, mT0, and BLOOMZ}
\begin{figure*}[ht!]
    \centering
    \includegraphics[width=0.69\linewidth]{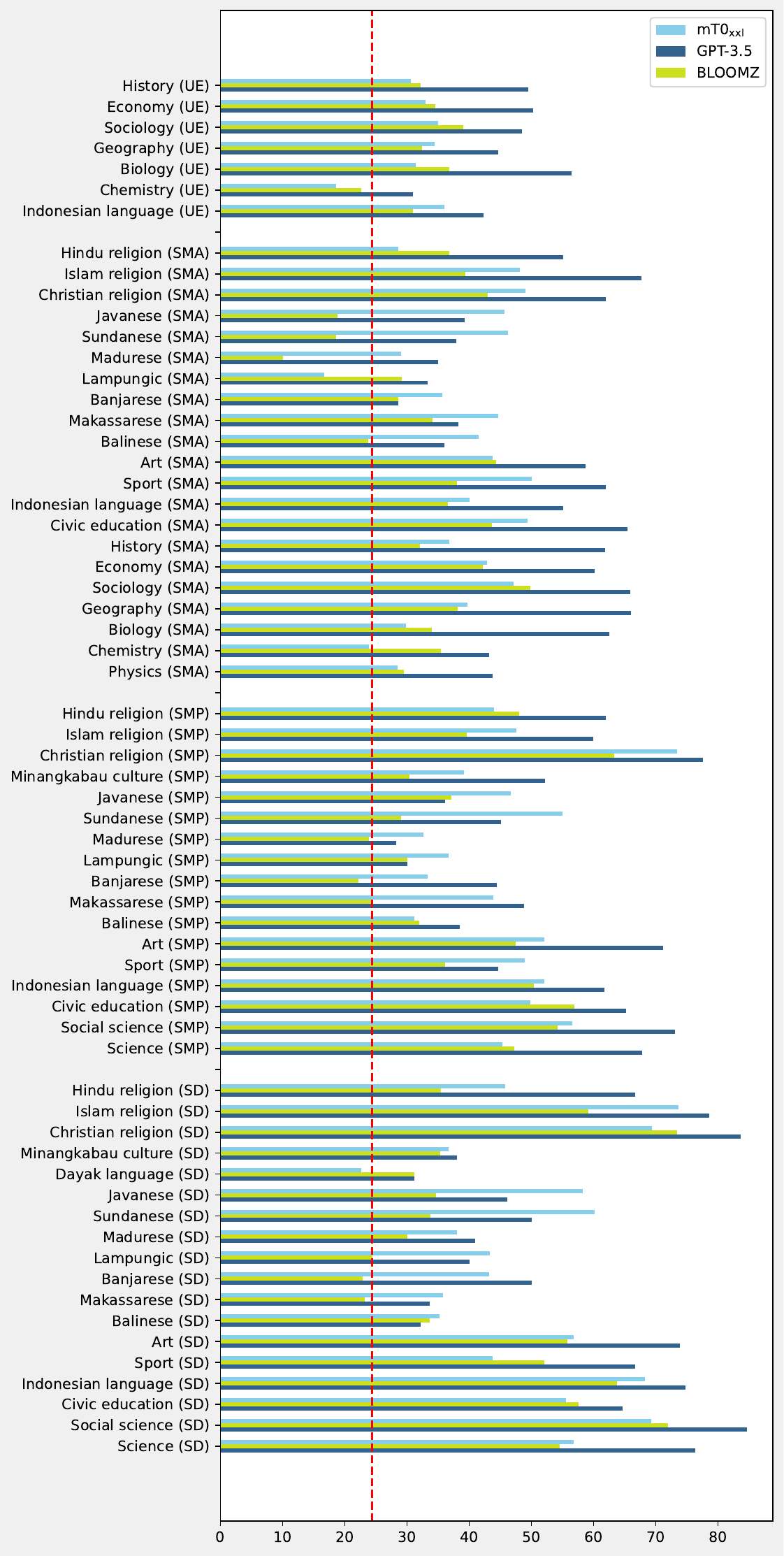} 
    \caption{Performance (\% accuracy) breakdown across the 64 tasks. ``SD'', ``SMP'', ``SMA'', ``UE'' indicate primary school, junior high school, senior high school, and university entrance tests, respectively. The red vertical line denotes random performance.}
    \label{fig:all}
\end{figure*}

\end{document}